\documentclass[accepted]{article}

\usepackage[utf8]{inputenc}
\usepackage[T1]{fontenc}
\usepackage[pdfusetitle]{hyperref}
\hypersetup{
  pdftitle={Temporal Causal Prior-Data Fitted Networks for Panel Data with Learned Reliability Signals},
  pdfauthor={Shravan Talupula, Saurabh Sharma},
  pdfsubject={Causal inference; foundation models; time series},
  pdfkeywords={causal discovery, prior-data fitted networks, time series, zero-shot}
}
\usepackage{url}
\usepackage{booktabs}
\usepackage{amsfonts}
\usepackage{amsmath}
\usepackage{amssymb}
\usepackage{nicefrac}
\usepackage{microtype}
\usepackage{graphicx}
\usepackage{xcolor}
\usepackage{subcaption}
\usepackage{multirow}
\usepackage{enumitem}
\usepackage{float}
\usepackage[numbers,compress]{natbib}

\title{Temporal Causal Prior-Data Fitted Networks\\for Panel Data with Learned Reliability Signals}

\author{
  Shravan Talupula, Saurabh Sharma\\
  ProfitOps Inc.\\
  \texttt{\{stalupula, ssharma\}@profitops.ai}
}

\date{}

\begin{document}
\maketitle
% ============================================================================
% Abstract
% ============================================================================
\begin{abstract}
Estimating causal effects in industrial time series -- such as manufacturing sensor data -- requires handling temporal dynamics, time-varying treatments, unobserved confounders, and streaming data. Established causal foundation models operate on static cross-sectional data (CausalPFN, CausalFM); neural temporal methods (CRN, G-Net) require per-dataset training; and concurrent temporal-PFN proposals to date (CausalTimePrior~\cite{thumm2026causaltimeprior}, the continuous-time Neural-ODE approach~\cite{thumm2026continuous}) are either prior-design work validated on synthetic samples only or have not been demonstrated at industrial scale. No existing causal PFN method outputs explicit per-pair reliability signals -- null-effect probability, identifiability, mediation, regime classification -- alongside its CATE estimates.

We introduce \textbf{Temporal Causal Prior-Data Fitted Networks (TCPFN)}, a foundation model for zero-shot temporal causal discovery with learned reliability signals -- applied to unseen time series without per-dataset training.

TCPFN makes four contributions. First, we introduce a \textbf{Causal Judgment Head} -- a structured output module that jointly predicts null-effect probability, confounding strength, identifiability, mediation fraction, and causal regime for each treatment-outcome pair. This enables the model to not only estimate effects but also flag when those estimates may be unreliable. The judgment head is trained jointly with the CATE objective using a three-phase curriculum that prevents destructive interference between tasks. Second, we train with a \textbf{mixed prior} that exposes the model to diverse causal structures during pretraining: a Causal Regime Prior covering independent (T $\perp\!\!\!\perp$ Y), direct, confounded (true effect = 0), mediated, time-varying confounded, and feedback regimes, combined with CausalFM-style identification priors for front-door and instrumental-variable structures. Each regime provides ground-truth supervision for all judgment head outputs, ensuring the model sees null-effect data during training -- critical for learning to distinguish causation from correlation. Third, a \textbf{discrete-token panel-data architecture} brings the PFN paradigm to temporal causal inference: each (unit, timestep) becomes a token with a three-part encoding (relative time from treatment, phase indicator, elapsed time for irregular sampling), and a cross-attention-only architecture prevents information leakage between prediction horizons. The discrete-token design complements concurrent continuous-time approaches~\cite{thumm2026continuous} and fits naturally for regularly-sampled industrial sensor data. Fourth, \textbf{zero-shot inference at industrial scale} via FAISS-based context selection and one-step posterior correction (OSPC) for valid confidence intervals, enabling V=1{,}275 causal discovery in a single GPU pass.

For causal discovery, TCPFN estimates the temporal CATE (conditional average treatment effect) for each variable pair and computes an \emph{edge score} that combines effect magnitude with null-probability and identifiability from the judgment head: $\text{score}(i \to j) = |\hat{\tau}_{i \to j}| \times (1 - p_{\text{null}}) \times p_{\text{identifiable}}$.

We evaluate on 19 zero-shot benchmark datasets across five domains (linear, nonlinear, chaotic, industrial, biological). On the Tennessee Eastman industrial benchmark (52 variables), TCPFN achieves AUROC 0.96 zero-shot -- approaching PCMCI's 1.00 without per-dataset fitting. On SWaT (water treatment, 51 variables) TCPFN reaches AUROC 0.93; on Causal Rivers (hydrological network), 0.98; on CAUSRCA (Fraunhofer CNC lathe, 104 variables), 0.97 -- and with 160 predicted edges vs.\ Granger's 6{,}164 on a graph with 104 true edges, TCPFN returns a graph an operator can plausibly inspect rather than a saturated adjacency matrix. On the nonlinear CauseMe NVAR-5 benchmark, TCPFN reaches AUROC 0.96 with best F1 0.80, versus Granger's best F1 0.67. On the Sachs protein signaling benchmark (a non-temporal flow-cytometry dataset treated as pseudo-timeseries), TCPFN's AUROC of 0.63 still exceeds both Granger (0.48) and PCMCI (0.46). The judgment head reliably detects null effects (NullF1 0.94, AUROC 0.99, Brier 0.04) with strong separation between null and non-null batches (NullSep 0.86), enabled by an ``independent'' training regime that teaches the model what ``no causal effect'' looks like. Per-token CATE trajectory \emph{shape} remains a limitation (trajectory correlation near zero on average), though aggregated effect magnitudes used for discovery are robust and effect-shape PEHE is 0.40--0.60 across the six tested shapes. We emphasize that the judgment head's outputs are learned heuristics, not formal guarantees.

On a proprietary Kraft pulp-and-paper manufacturing dataset, TCPFN completes V=1,275 causal discovery in 6 hours on a single H100 NVL GPU, zero-shot (the V=1,275 industrial discovery run is on H100 NVL; the model itself was pretrained on a single RTX 5090 in $\sim$4 hours, see Section~\ref{sec:industrial_eval} and Appendix~\ref{app:training}). The discovered graph's structure reflects the plant's continuous-flow production layout, with edge density concentrated around the paper machine and decreasing through the upstream stages. Against PCMCI on the same data, TCPFN's top edges concentrate on cross-subsystem causal relationships across plant areas; PCMCI's top edges concentrate on within-instrument controller-measurement coupling at lag~1 (relationships already documented on the plant's P\&ID schematics). We treat the Kraft deployment as a scalability and hypothesis-generation case study rather than as quantitative validation -- the dataset has no ground-truth causal graph and operator-side validation of specific candidate edges is outside the scope of this paper.
\end{abstract}

% ============================================================================
% 1. Introduction
% ============================================================================
\section{Introduction}
\label{sec:introduction}

\subsection{Motivation}

Modern manufacturing generates massive multivariate time series from sensors, actuators, and quality metrics. Plant operators routinely face causal questions: \emph{``What happens to product quality if I increase flow rate?''} or \emph{``Which sensor is causing this defect?''} Answering such questions requires causal inference -- not mere correlation -- yet classical approaches such as randomized controlled trials are infeasible in continuous production environments.

Observational causal inference from time series is the only viable option, but existing methods require per-dataset model training, which is impractical when each production line has unique dynamics and operators need answers in real time.

\subsection{Limitations of Existing Work}

Prior-Data Fitted Networks (PFNs)~\cite{muller2022transformers} have emerged as a powerful paradigm for amortized inference: pretrain on synthetic data drawn from a prior, then apply zero-shot to real data. This paradigm has been successfully applied to tabular prediction (TabPFN~\cite{hollmann2023tabpfn}), static causal inference (CausalPFN~\cite{balazadeh2024causalpfn}), and extended identification strategies (CausalFM~\cite{causalfm2026}). Concurrent work explores temporal extensions through synthetic prior design (CausalTimePrior~\cite{thumm2026causaltimeprior}, a workshop-scale data generator) and continuous-time Neural-ODE architectures~\cite{thumm2026continuous}. However, the established (scaled) causal PFN models all operate on \emph{static, cross-sectional data} and cannot handle:

\begin{itemize}[nosep,leftmargin=*]
    \item \textbf{Temporal dynamics}: treatment effects that unfold over time (delayed, oscillating, decaying)
    \item \textbf{Time-varying treatments}: repeated interventions, not just one-time binary treatments
    \item \textbf{Irregular sampling}: missing observations and variable time gaps between measurements
    \item \textbf{Streaming data}: real-time updates as new sensor readings arrive
\end{itemize}

Neural temporal causal methods (RMSN~\cite{lim2018forecasting}, CRN~\cite{bica2020estimating}, G-Net~\cite{li2021gnet}) address temporal dynamics but require per-dataset training, limiting their applicability to settings with abundant labeled data and sufficient training time.

Furthermore, no existing causal PFN method -- including the concurrent temporal work cited above -- outputs explicit per-pair reliability signals (null-effect probability, identifiability, mediation, regime classification) alongside its CATE estimates. Classical sensitivity analysis methods exist but require explicit assumptions and per-dataset analysis.

\subsection{Contributions}

We introduce \textbf{Temporal Causal Prior-Data Fitted Networks (TCPFN)}, a foundation model for temporal causal inference on industrial time series. We make four contributions:

\begin{enumerate}[nosep,leftmargin=*]
    \item \textbf{Temporal token design for causal PFNs (Section~\ref{sec:temporal_tokens}).} We extend Prior-Data Fitted Networks from static/cross-sectional to panel data. Each (unit, timestep) pair becomes a token with a three-part encoding (relative time from treatment, phase indicator, elapsed time for irregular sampling). A cross-attention-only architecture prevents information leakage between prediction horizons. The discrete-token design complements concurrent continuous-time approaches~\cite{thumm2026continuous}: discrete tokens fit naturally for sampled-sensor data (DCS/SCADA at fixed intervals), while continuous-time formulations target irregular clinical data. The architecture captures delayed, oscillating, and decaying treatment effects that static methods cannot.

    \item \textbf{Causal Judgment Head (Section~\ref{sec:judgment_head}).} We introduce a structured output module that jointly predicts null-effect probability, confounding strength, identifiability, mediation fraction, and causal regime. This enables the model to not only estimate effects but also to \emph{flag when those estimates may be unreliable} -- a capability not present in existing causal PFN methods. The judgment head is trained jointly with the CATE objective using a three-phase curriculum that prevents the auxiliary task from destabilizing effect estimation: CATE-only $\to$ null detection $\to$ full causal reasoning.

    \item \textbf{Causal Regime Prior (Section~\ref{sec:prior}).} We train on a mixed prior: a Causal Regime Prior covering independent (T $\perp\!\!\!\perp$ Y), direct, confounded (true effect = 0), mediated, time-varying confounded, and feedback regimes, combined with CausalFM-style identification priors for front-door and instrumental-variable structures. Each regime provides ground-truth metadata for all judgment head outputs. This teaches the model to distinguish causation from correlation -- unlike standard priors that only generate direct effects with varying confounding.

    \item \textbf{Zero-shot inference at scale (Section~\ref{sec:inference}).} We pair the temporal causal model with FAISS-based context selection and one-step posterior correction (OSPC) for valid confidence intervals, enabling V=1{,}275 causal discovery in a single GPU pass with no per-dataset fitting.
\end{enumerate}

A key insight is that the \emph{same} temporal CATE model performs causal discovery through interventional reasoning: for each variable pair, estimating ``what happens to $Y$ when I intervene on $X$?'' The judgment head further improves discovery by down-weighting edges it judges as confounded or unidentifiable: $\text{score}(i \to j) = |\hat{\tau}| \times (1 - p_{\text{null}}) \times p_{\text{id}}$.

We evaluate on 19 benchmark datasets across five domains (linear, nonlinear, chaotic, industrial, biological) and on a proprietary Kraft pulp-and-paper manufacturing dataset, where TCPFN completes V=1,275 zero-shot causal discovery in 6 hours on a single H100 NVL GPU (preprocessing details in Section~\ref{sec:industrial_eval}; pretraining itself ran on a single RTX 5090, see Appendix~\ref{app:training}). The discovered graph's structure reflects the plant's continuous-flow production layout, with edge density concentrated around the paper machine. Compared head-to-head against PCMCI on the same dataset, TCPFN's top edges identify cross-subsystem causal relationships; PCMCI's top edges are dominated by trivial within-instrument coupling already documented on the plant's P\&ID schematics -- a distinction that matters when the goal is to surface new causal structure rather than rediscover known control loops.

% ============================================================================
% 2. Background
% ============================================================================
\section{Background}
\label{sec:background}

\subsection{Prior-Data Fitted Networks}

Prior-Data Fitted Networks (PFNs)~\cite{muller2022transformers} amortize Bayesian inference by pretraining a transformer on synthetic datasets drawn from a prior $p(\mathcal{D})$. At inference time, the model performs approximate posterior predictive inference in a single forward pass, without gradient updates. This paradigm has been applied to tabular classification (TabPFN~\cite{hollmann2023tabpfn}), regression, and static causal inference (CausalPFN~\cite{balazadeh2024causalpfn}).

CausalPFN trains on synthetic data from structural causal models (SCMs) where ground-truth potential outcomes $Y(0), Y(1)$ are known. The model learns to estimate conditional average treatment effects (CATE) $\tau(x) = \mathbb{E}[Y(1) - Y(0) \mid X = x]$ from observational data, using in-context learning. CausalFM~\cite{causalfm2026} extends this to front-door adjustment and instrumental variable settings.

\textbf{Limitations.} All existing causal PFNs (1) operate on static, cross-sectional data -- each unit $i$ has a single outcome $Y_i$, not a trajectory $\{Y_i(t)\}_{t=1}^T$ -- and (2) output only a point estimate $\hat{\tau}(x)$ with no assessment of \emph{whether the estimate is trustworthy}: they cannot distinguish genuine causal effects from spurious confounded associations, nor flag when effects are unidentifiable from the observed data.

\subsection{Temporal Causal Inference}

In the potential outcomes framework for longitudinal data~\cite{robins1986new}, we observe panel data $\{(X_i, T_i(t), Y_i(t))\}_{i=1}^N$ for $t = 1, \ldots, T$, where $X_i$ are static covariates, $T_i(t)$ is the treatment at time $t$, and $Y_i(t)$ is the outcome. The temporal CATE is:
\begin{equation}
    \tau(x, t) = \mathbb{E}[Y_i(t) \mid \text{do}(T_i(t_0) = 1), X_i = x] - \mathbb{E}[Y_i(t) \mid \text{do}(T_i(t_0) = 0), X_i = x]
\end{equation}

Classical methods include g-computation~\cite{robins1986new}, marginal structural models (MSMs)~\cite{robins2000marginal}, and structural nested models. Neural approaches include RMSN~\cite{lim2018forecasting}, Counterfactual Recurrent Networks (CRN)~\cite{bica2020estimating}, and G-Net~\cite{li2021gnet}. All require per-dataset training.

\textbf{Key challenge.} Time-varying confounding with treatment-confounder feedback: past outcomes influence future treatment assignment, violating standard exchangeability assumptions.

\subsection{Causal Discovery from Time Series}

Temporal causal discovery aims to recover a directed acyclic graph (DAG) $\mathcal{G}$ with adjacency matrix $A \in \{0,1\}^{V \times V \times L}$, where $A_{j,k,l} = 1$ indicates that variable $j$ causes variable $k$ at lag $l$. Methods range from Granger causality~\cite{granger1969investigating} to constraint-based approaches (PCMCI~\cite{runge2019detecting}) and continuous optimization (DYNOTEARS~\cite{pamfil2020dynotears}).

\textbf{Gap.} No existing method provides both amortized (zero-shot) causal discovery \emph{and} treatment effect estimation in a unified system, nor can any method assess the identifiability or reliability of its own causal conclusions.

% ============================================================================
% 3. Method
% ============================================================================
\section{Method}
\label{sec:method}

\begin{figure}[t]
    \centering
    \includegraphics[width=\textwidth]{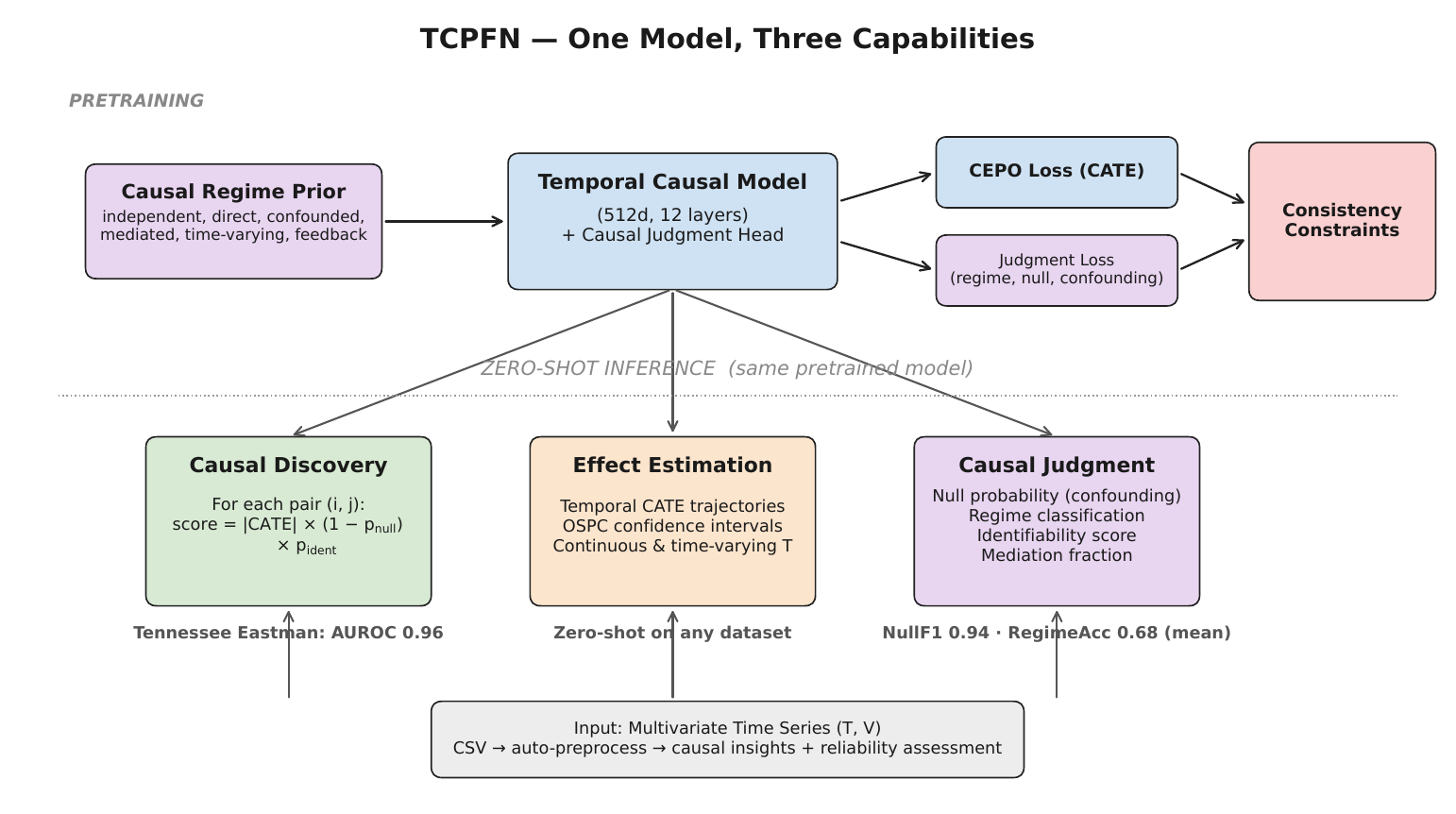}
    \caption{\textbf{TCPFN system overview.} A single temporal causal model is pretrained on synthetic SCMs (top) and applied zero-shot to real data (bottom). The same model performs causal discovery (pairwise interventional CATE), temporal effect estimation (CATE trajectories), and causal judgment (learned reliability signals).}
    \label{fig:system_overview}
\end{figure}

Figure~\ref{fig:system_overview} presents the TCPFN system. A single temporal model, pretrained on synthetic SCMs, performs effect estimation and causal discovery zero-shot.

%  --  --  --  --  --  --  --  --  --  --  --  --  --  --  --  --  --  --  --  --  --  --  --  --  -- -
\subsection{Problem Setup}
\label{sec:problem_setup}

We observe panel data $\{(X_i, T_i(t), Y_i(t))\}_{i=1}^N$ for $t = 1, \ldots, T_{\text{pre}} + T_{\text{post}}$, where $X_i \in \mathbb{R}^D$ are static covariates, $T_i(t) \in \{0,1\}$ (or $\mathbb{R}$ for continuous treatments) is the treatment status, and $Y_i(t) \in \mathbb{R}^K$ are outcome variables.

\textbf{Goal 1 (Estimation).} Estimate the temporal CATE:
\begin{equation}
    \tau(x, h) = \mathbb{E}[Y(t_0 + h) \mid \text{do}(T(t_0) = 1), X = x] - \mathbb{E}[Y(t_0 + h) \mid \text{do}(T(t_0) = 0), X = x]
\end{equation}
for each post-treatment horizon $h = 0, 1, \ldots, T_{\text{post}} - 1$.

\textbf{Goal 2 (Discovery).} Recover the temporal causal adjacency matrix $A \in [0,1]^{V \times V \times L}$, where $A_{j,k,l}$ indicates the strength of the causal effect of variable $j$ on variable $k$ at lag $l$.

%  --  --  --  --  --  --  --  --  --  --  --  --  --  --  --  --  --  --  --  --  --  --  --  --  -- -
\subsection{Temporal Token Design}
\label{sec:temporal_tokens}

The core architectural innovation of TCPFN is representing panel data as a sequence of \emph{temporal tokens}, enabling in-context causal inference over time.

\textbf{Tokenization.} Each (unit $i$, timestep $t$) pair becomes one token with features:
\begin{equation}
    \text{token}_{i,t} = [T_i, \; X_i, \; Y_{\text{pre},i}(1{:}T_{\text{pre}}), \; \mathbf{0}_{\text{pad}}] \in \mathbb{R}^{F_{\max}}
\end{equation}
where $T_i$ is the treatment indicator, $X_i$ are static covariates, and $Y_{\text{pre},i}$ is the pre-treatment trajectory, zero-padded to $F_{\max}$ features.

\textbf{Temporal Encoding.} Each token receives a three-part additive encoding:
\begin{equation}
    \text{enc}(i, t) = \underbrace{E_{\text{rel}}(\delta_t)}_{\text{relative time}} + \underbrace{E_{\text{phase}}(\phi_t)}_{\text{pre/post}} + \underbrace{W_{\text{elapsed}} \cdot \Delta t}_{\text{irregular gaps}}
\end{equation}
where $\delta_t = t - t_0$ is the relative time from treatment, $\phi_t = \mathbb{1}[t \geq t_0]$ is the phase indicator, $E_{\text{rel}} \in \mathbb{R}^{(2H+1) \times d}$ and $E_{\text{phase}} \in \mathbb{R}^{2 \times d}$ are learnable embeddings, $W_{\text{elapsed}} \in \mathbb{R}^{1 \times d}$ projects the actual elapsed time $\Delta t$ between observations (supporting irregular sampling), and all are initialized with small standard deviation (0.02) so the model starts close to static behavior.

\textbf{Cross-Attention Only.} The key design choice is that \emph{query tokens never attend to other query tokens} -- only to context tokens. This prevents information leakage between prediction horizons and ensures each horizon $h$ predicts independently:
\begin{equation}
    \text{Attn}(Q, K, V) = \text{softmax}\!\left(\frac{Q_{\text{query}} K_{\text{context}}^\top}{\sqrt{d}}\right) V_{\text{context}}
\end{equation}
Context tokens (full trajectories of training units) attend to each other and to themselves. Query tokens (post-treatment timesteps of test units) attend only to the context.

\begin{figure}[t]
    \centering
    \includegraphics[width=\textwidth]{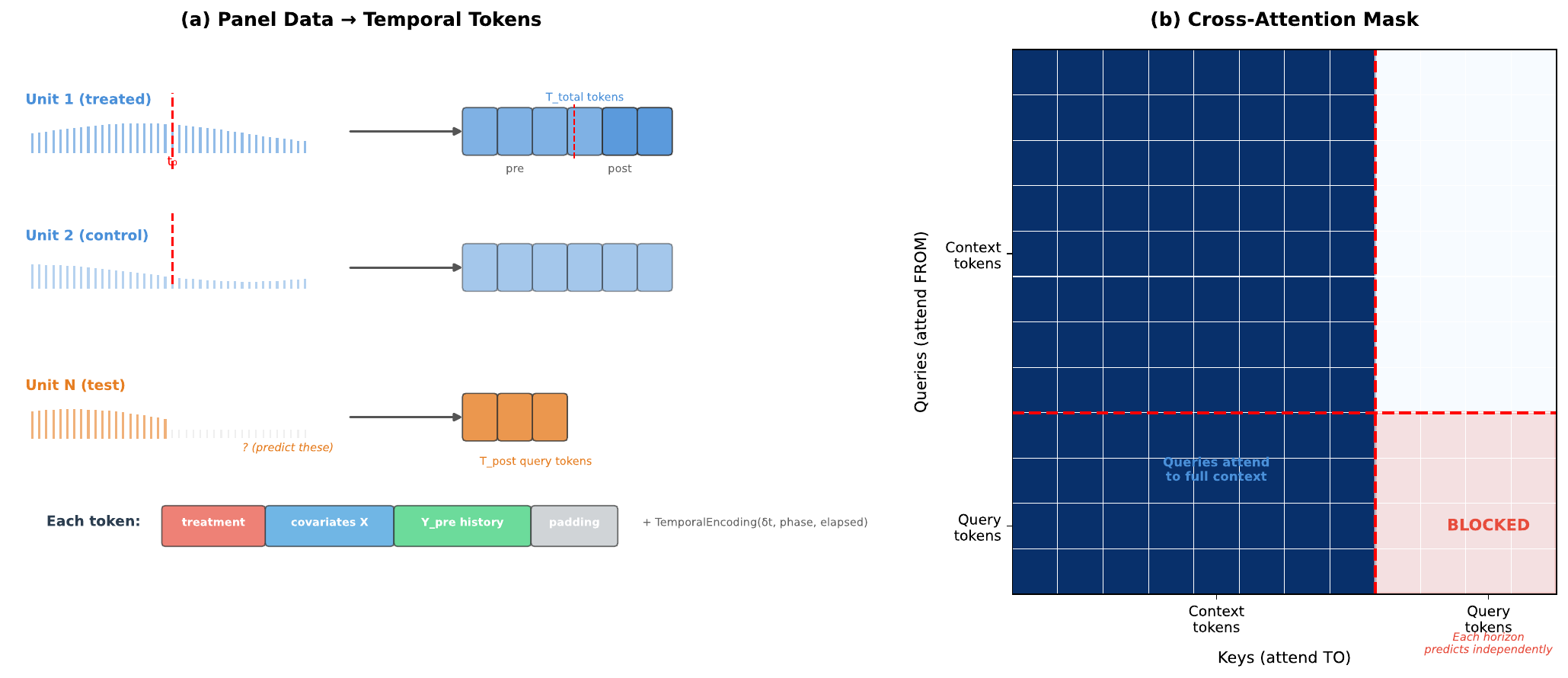}
    \caption{\textbf{Temporal token design.} (a)~Panel data is converted to temporal tokens: each (unit, timestep) pair becomes one token. Context units have full trajectories; query units have only post-treatment tokens. (b)~Cross-attention mask: queries attend to the full context but \textbf{never to each other} (red BLOCKED region), preventing information leakage between horizons.}
    \label{fig:temporal_tokens}
\end{figure}

\begin{figure}[t]
    \centering
    \includegraphics[width=\textwidth]{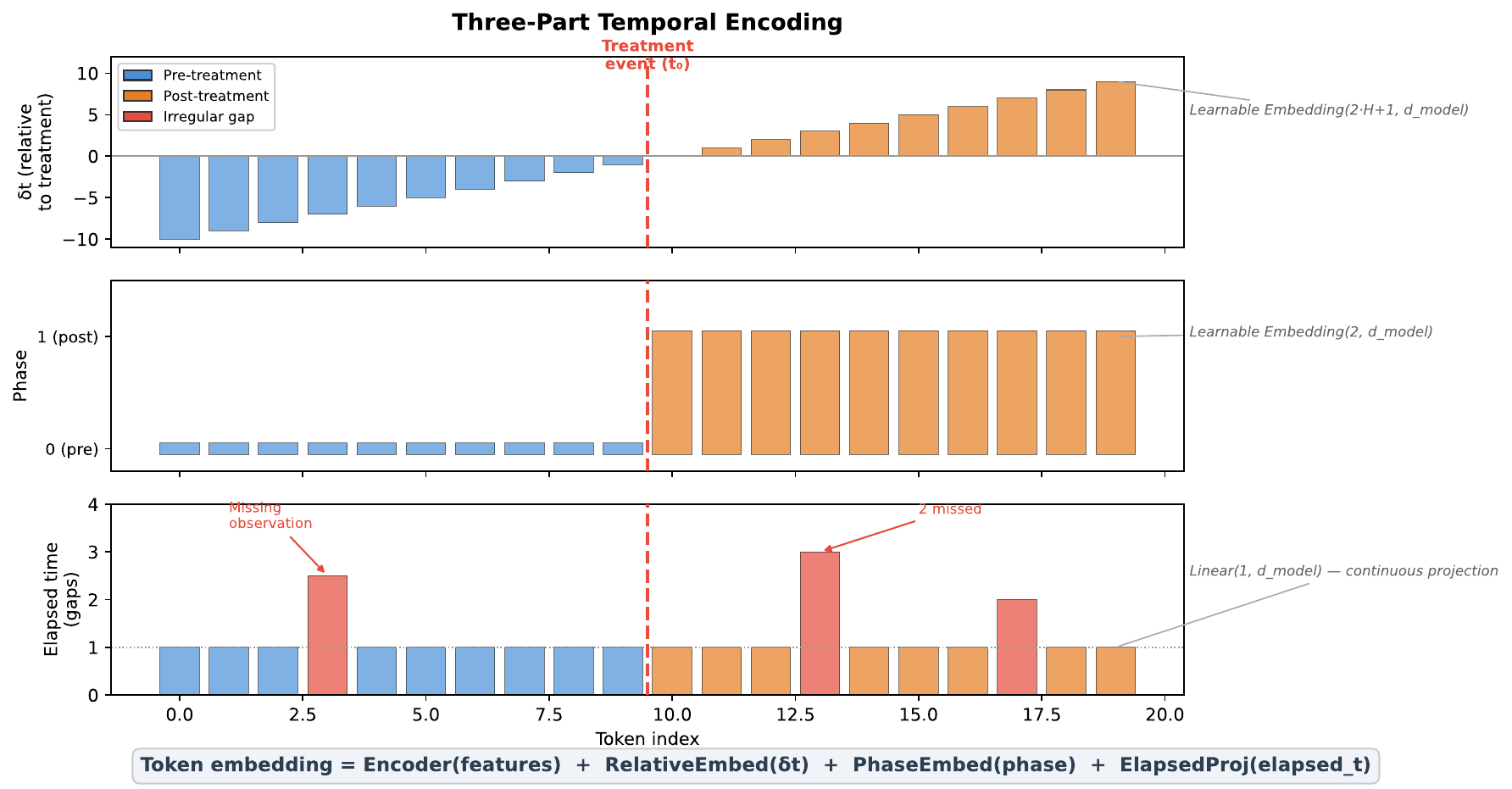}
    \caption{\textbf{Three-part temporal encoding.} Top: relative time from treatment ($\delta_t$), encoded via learnable embedding. Middle: binary phase indicator (pre/post treatment). Bottom: elapsed time between observations -- regular sampling yields 1.0; gaps from missing observations yield larger values, projected via a learned linear layer. The combined encoding is added to token embeddings.}
    \label{fig:temporal_encoding}
\end{figure}

%  --  --  --  --  --  --  --  --  --  --  --  --  --  --  --  --  --  --  --  --  --  --  --  --  -- -
\subsection{Model Architecture}
\label{sec:architecture}

\subsubsection{Treatment Effect Model}

The \textsc{TemporalTabDPTModel} processes temporal token sequences:

\begin{enumerate}[nosep,leftmargin=*]
    \item \textbf{Feature encoder:} $\text{LayerNorm}(\text{Linear}(F_{\max}, d))$
    \item \textbf{Outcome encoder:} $\text{LayerNorm}(\text{Linear}(K, d))$ for $K$ outcome dimensions
    \item \textbf{Temporal encoding:} Added to all token embeddings (Section~\ref{sec:temporal_tokens})
    \item \textbf{Context fusion:} Context tokens combine feature and outcome encodings; query tokens use features only
    \item \textbf{Transformer:} $L$ layers of cross-attention (Section~\ref{sec:temporal_tokens})
    \item \textbf{Output head:} $\text{Linear}(d, d_{\text{ff}}) \to \text{GELU} \to \text{Linear}(d_{\text{ff}}, K \times B)$, producing $B$-bin histogram logits per outcome per query token
\end{enumerate}

\textbf{Training loss.} We use the HL-Gaussian cross-entropy loss~\cite{imani2018improving} over 128-bin histograms with CEPO (Conditional Expected Potential Outcomes) standardization. For each query token, a random treatment $t \in \{0,1\}$ is assigned, and the loss compares predicted histograms against ground-truth $\mathbb{E}[Y(t)]$:
\begin{equation}
    \mathcal{L} = -\sum_{b=1}^{B} p_b^{\text{target}} \log \hat{p}_b, \quad p_b^{\text{target}} = \Phi\!\left(\frac{u_b - y}{\sigma}\right) - \Phi\!\left(\frac{l_b - y}{\sigma}\right)
\end{equation}
where $l_b, u_b$ are bin edges, $y$ is the Z-standardized ground-truth outcome, $\Phi$ is the standard normal CDF, and $\sigma = 0.5$ controls smoothing.

\subsubsection{Causal Discovery via Interventional Reasoning}

A key insight of TCPFN is that the \emph{same} temporal CATE model used for effect estimation can perform causal discovery. Rather than learning a separate discovery model -- which we found collapses to predicting base rates regardless of architecture -- we use the treatment effect model to answer: ``does intervening on variable $i$ change the outcome of variable $j$?''

\textbf{Pairwise CATE estimation.} For each variable pair $(i, j)$:
\begin{enumerate}[nosep,leftmargin=*]
    \item Construct panel data from sliding windows of the multivariate time series
    \item Binarize treatment: $T_i = \mathbb{1}[X_i > \text{median}(X_i)]$ in the pre-treatment window
    \item Include other variables as covariates: $[X_k^{\text{mean}}, X_k^{\text{std}}, X_k^{\text{trend}}]$ for $k \neq i, j$ -- this enables confounding adjustment
    \item Estimate $\hat{\tau}(i \to j) = \hat{\mathbb{E}}[Y_j \mid \text{do}(T_i{=}1)] - \hat{\mathbb{E}}[Y_j \mid \text{do}(T_i{=}0)]$ using the temporal CATE model
    \item Edge score: $A_{i,j} = |\hat{\tau}(i \to j)|$
\end{enumerate}

This is a pairwise CATE-based discovery heuristic: the model estimates what happens under hypothetical treatment changes, rather than testing correlations. While this is not formal interventional reasoning (the model operates on observational data with covariate adjustment, not true do-calculus), it leverages the PFN's pretraining on synthetic interventional data to approximate interventional effects. Because the model was pretrained on diverse SCMs where ground-truth interventional effects are known, it has learned to distinguish direct causal effects from spurious correlations -- the same covariates that the model uses for confounding adjustment in effect estimation serve as conditioning variables that filter out indirect effects.

\textbf{Edge scoring with causal judgment.} When the Causal Judgment Head (\S\ref{sec:judgment_head}) is available, the edge score incorporates the model's confidence in the causal relationship:
\begin{equation}
    A_{i,j} = |\hat{\tau}(i \to j)| \times (1 - p_{\text{null}}) \times p_{\text{identifiable}}
\end{equation}
where $p_{\text{null}}$ is the predicted probability that the true effect is zero (confounding detection) and $p_{\text{identifiable}}$ is the predicted probability that the effect is identifiable from the observed data. This down-weights edges the model judges as confounded or unidentifiable -- a capability not present in existing scaled causal PFN methods, and, from the public descriptions of concurrent temporal-PFN work~\cite{thumm2026causaltimeprior, thumm2026continuous}, not provided by them either.

\textbf{Computational cost.} Pairwise estimation requires $O(V^2)$ model calls, which is slower than Granger or PCMCI for large $V$. However, each call is a single forward pass through a small transformer (259\,MB), and pairs can be parallelized across multiple GPUs.

%  --  --  --  --  --  --  --  --  --  --  --  --  --  --  --  --  --  --  --  --  --  --  --  --  -- -
\subsection{Causal Judgment Head}
\label{sec:judgment_head}

A causal foundation model should not only predict effects -- it should infer \emph{when} effects are identifiable, mediated, spurious, or insufficient. We introduce the \textbf{Causal Judgment Head}, a structured output module that maps transformer hidden states to a full causal judgment for each treatment-outcome pair.

\subsubsection{Architecture}

Figure~\ref{fig:judgment_head} illustrates the architecture. The judgment head takes the transformer's final hidden states $H \in \mathbb{R}^{B \times L_q \times d}$ (pooled across query tokens) and produces:

\begin{figure}[t]
    \centering
    \includegraphics[width=\textwidth]{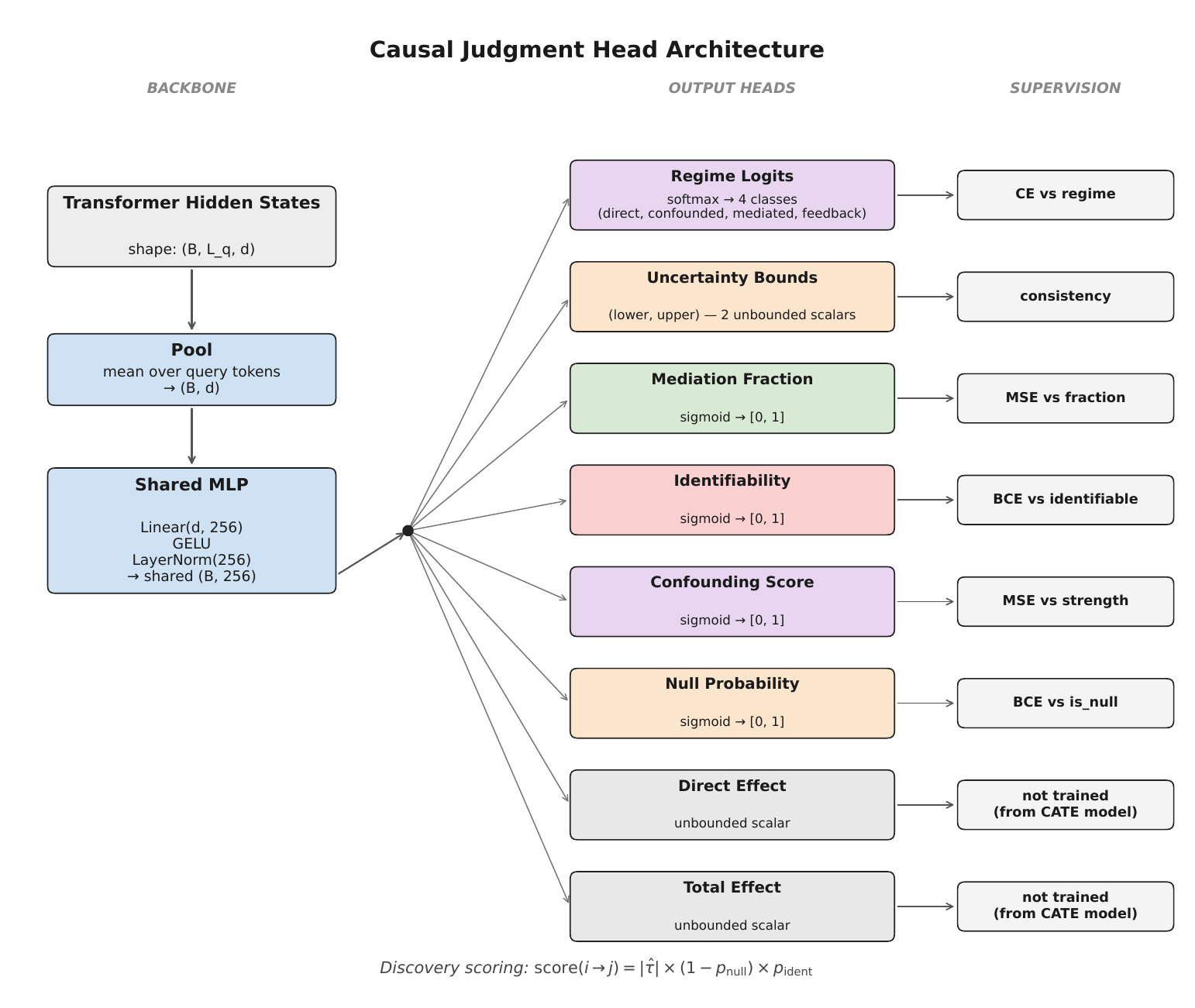}
    \caption{\textbf{Causal Judgment Head architecture.} Transformer hidden states are pooled, passed through a shared MLP (Linear $\to$ GELU $\to$ LayerNorm), then routed to 8 output heads: 5 supervised task heads (Regime, Mediation, Identifiability, Confounding, Null), 1 consistency-bound head (Uncertainty Bounds), and 2 untrained effect heads (Direct/Total Effect, rendered greyed-out; effect magnitudes are obtained from the main CATE model at inference time, not from the judgment head). Binary heads use BCE; regime uses cross-entropy; continuous heads use MSE.}
    \label{fig:judgment_head}
\end{figure}

The judgment head produces five supervised outputs (Figure~\ref{fig:judgment_head} also shows a sixth consistency-bound head, ``Uncertainty Bounds'' (lower, upper), trained against an internal consistency loss rather than against external ground truth; and two architectural slots for Direct/Total Effect magnitudes that are not trained by the judgment objective and are populated at inference from the CATE model):

\begin{enumerate}[nosep,leftmargin=*]
    \item \textbf{Null probability} $p_{\text{null}} \in [0,1]$: $P(\text{true effect} = 0 \mid \text{data})$ -- confounding detector
    \item \textbf{Confounding score} $c \in [0,1]$: estimated strength of unmeasured confounding
    \item \textbf{Identifiability} $p_{\text{id}} \in [0,1]$: $P(\text{effect is identifiable from observed data})$
    \item \textbf{Mediation fraction} $m \in [0,1]$: fraction of total effect mediated by observed variables
    \item \textbf{Causal regime}: $\arg\max$ over four regime classes $\{\text{direct, confounded, mediated, feedback}\}$. The prior generates six regimes (Section~\ref{sec:prior}); for the 4-way regime cross-entropy loss, the two regimes without a dedicated output class -- Independent and Time-varying confounded -- are folded into the \emph{confounded} class, so the classifier sees the consolidated labels \{direct, confounded $\cup$ independent $\cup$ time-varying, mediated, feedback\}. This consolidation is structural (it groups regimes that share the strongest visible association pattern), not semantic: Independent is a null-effect regime, but Time-varying confounded is not. The null-vs-non-null distinction is carried separately by the binary null-probability head.
\end{enumerate}

Effect magnitudes ($\hat{\tau}_{\text{total}}$, $\hat{\tau}_{\text{direct}}$) are obtained from the main CATE model at inference time via \texttt{predict\_cepo\_temporal()}, not from the judgment head -- the CATE model is specifically optimized for effect estimation and produces superior predictions.

\textbf{Important caveat.} These outputs are \emph{learned heuristics} -- correlations between data patterns and prior-assigned labels. Identifiability is a property of the causal model and assumptions~\cite{pearl2009causality}, not something determinable from data patterns alone. The model learns to recognize patterns that \emph{correlated} with identifiability in the training prior. We do not claim formal identifiability analysis; see \S\ref{sec:judgment_analysis} for an honest assessment of what the model does and does not learn.

\subsubsection{Supervision}

Each judgment output is supervised with ground truth from the training prior:
\begin{itemize}[nosep,leftmargin=*]
    \item Null probability: binary cross-entropy against $\mathbb{1}[\text{is\_null\_effect}]$
    \item Regime: cross-entropy against the regime label from the prior
    \item Identifiability: binary cross-entropy against $\mathbb{1}[\text{is\_identifiable}]$ (matching the binary target shown in Figure~\ref{fig:judgment_head})
    \item Mediation fraction, confounding strength: MSE against prior metadata
\end{itemize}

The judgment loss is the \emph{average} across all task losses (not the sum), so each task contributes equally regardless of how many are active:
\begin{equation}
    \mathcal{L}_{\text{judg}} = \frac{1}{|\mathcal{T}|} \sum_{t \in \mathcal{T}} \ell_t
\end{equation}

For non-confounded batches, the total loss combines CATE and judgment:
\begin{equation}
    \mathcal{L} = \mathcal{L}_{\text{CATE}} + \lambda \mathcal{L}_{\text{judg}}, \quad \lambda = 0.1
\end{equation}

For confounded batches (where Z-standardization corrupts CATE targets due to confounding-induced group mean differences), only the judgment loss is used: $\mathcal{L} = \mathcal{L}_{\text{judg}}$.

\subsubsection{Curriculum Training Schedule}

Training the CATE objective and judgment head simultaneously creates destructive gradient interference: the judgment head's gradients destabilize the backbone before it has learned stable temporal representations. We address this with a three-phase curriculum:

\begin{enumerate}[nosep,leftmargin=*]
    \item \textbf{Phase 1 (steps 0--20K): CATE only.} The backbone learns stable temporal and effect representations without any judgment head interference. $\mathcal{L} = \mathcal{L}_{\text{CATE}}$.
    \item \textbf{Phase 2 (steps 20K--50K): Add null detection + confounding.} The easiest judgment tasks -- null-effect detection ($p_{\text{null}}$) and confounding strength -- are introduced. These binary/scalar tasks align well with CATE features since they reason about effect presence.
    \item \textbf{Phase 3 (steps 50K+): Add regime + identifiability + mediation.} The hardest tasks -- regime classification (4-class), identifiability, mediation fraction -- are enabled. By this point, the backbone has stable representations for both CATE and basic judgment, providing a foundation for structural causal reasoning.
\end{enumerate}

This curriculum is motivated by the observation that null-effect detection (binary: ``is there an effect?'') is learnable early, while regime classification (4-class: ``what kind of causal structure?'') requires richer representations that only emerge after extended CATE training. The phase boundaries (20K, 50K) are configurable hyperparameters.

%  --  --  --  --  --  --  --  --  --  --  --  --  --  --  --  --  --  --  --  --  --  --  --  --  -- -
\subsection{Training Prior Design}
\label{sec:prior}

TCPFN is pretrained on synthetic data generated on-the-fly from a diverse prior over temporal SCMs. Each training batch uses a freshly sampled SCM with randomized structure, ensuring the model generalizes across temporal dynamics.

\begin{figure}[t]
    \centering
    \includegraphics[width=\textwidth]{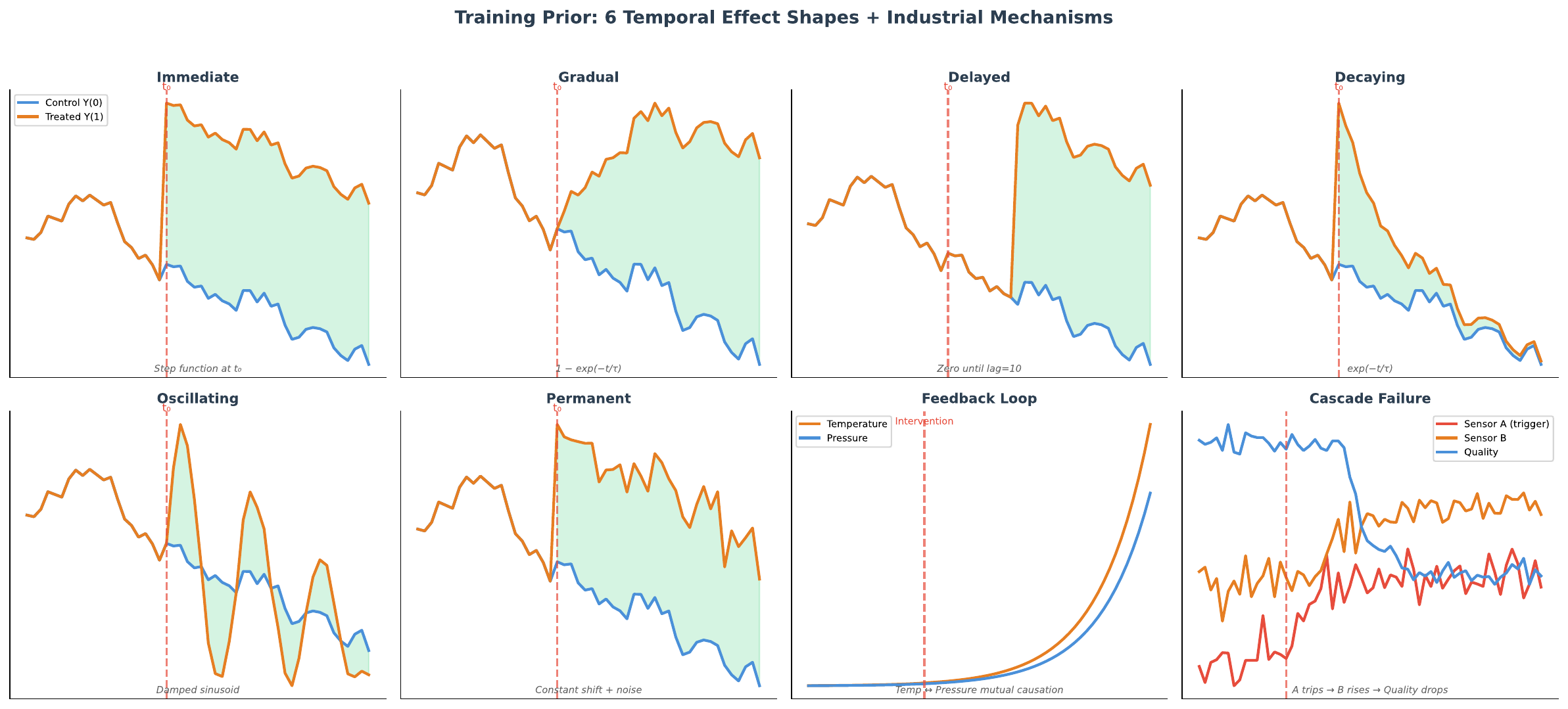}
    \caption{\textbf{Training prior diversity.} Top row: six temporal effect shapes generated by the base prior -- immediate, gradual, delayed, decaying, oscillating, and permanent. Bottom row: industrial mechanisms -- feedback loops (temperature$\leftrightarrow$pressure mutual causation) and cascade failures (sensor A triggers B, which degrades quality). Orange: treated trajectory; blue: control trajectory; green shading: treatment effect.}
    \label{fig:prior_effects}
\end{figure}

\subsubsection{Base Temporal Prior with Causal Regimes}

The base prior generates AR($p$) dynamics with six temporal effect shapes (immediate, gradual, delayed, decaying, oscillating, permanent) and diverse treatment assignment mechanisms.

Critically, the base prior samples from a \textbf{Causal Regime Prior} (Figure~\ref{fig:regime_prior}) that exposes the model to qualitatively different causal structures:

\begin{figure}[t]
    \centering
    \includegraphics[width=\textwidth]{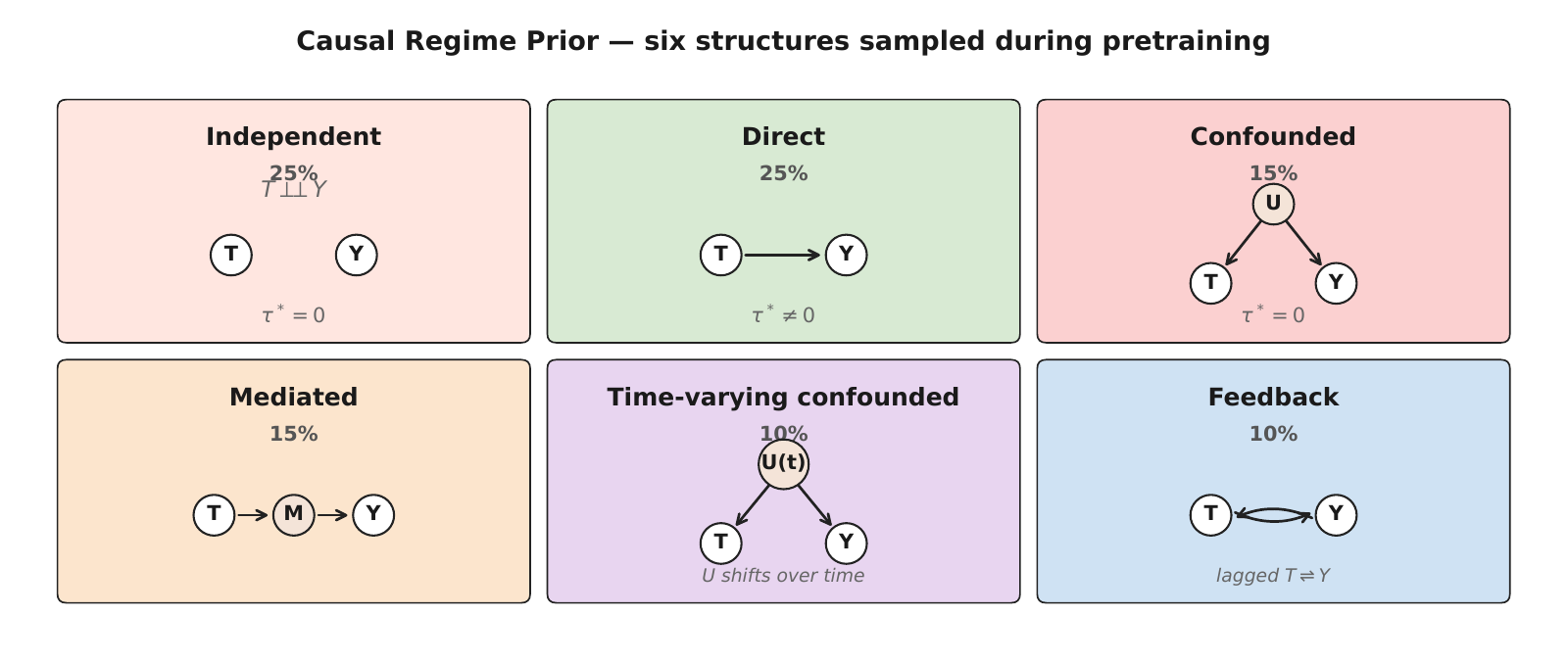}
    \caption{\textbf{Causal Regime Prior.} Six causal structures sampled during pretraining. Each regime provides ground-truth metadata for all judgment head outputs. The confounded (15\%) and independent (25\%) regimes both have zero true effect -- teaching the model to distinguish causation from correlation.}
    \label{fig:regime_prior}
\end{figure}
\begin{itemize}[nosep,leftmargin=*]
    \item \textbf{Independent (25\%):} $T \perp\!\!\!\perp Y$. Treatment is randomly assigned, no confounding, no causal effect. The most direct anti-hallucination signal: the model must learn to predict ``no effect'' when none exists.
    \item \textbf{Direct (25\%):} Standard treatment $\to$ outcome; confounding present but adjustable via covariates.
    \item \textbf{Confounded (15\%):} $T \leftarrow U \rightarrow Y$ with \emph{zero} true effect ($\tau^* = 0$). Strong association exists from shared causes, but the causal effect is null. Confounding strength $3\times$ amplified. The model must learn: \emph{strong association $\neq$ causal effect}.
    \item \textbf{Mediated (15\%):} $T \to M \to Y$. Effect flows entirely through a mediator variable.
    \item \textbf{Time-varying confounded (10\%):} Confounders change over time, requiring temporal adjustment.
    \item \textbf{Feedback (10\%):} $T \rightleftharpoons Y$. Treatment and outcome mutually cause each other across time lags.
\end{itemize}

Each regime generates ground-truth metadata for all judgment head outputs: $\texttt{is\_null\_effect}$, $\texttt{causal\_regime}$, $\texttt{identifiable}$, $\texttt{mediation\_fraction}$, and $\texttt{confounding\_strength}$.

\textbf{Design rationale.} Most causal PFN training priors generate only direct effects with varying confounding strength. By including null-effect data totaling 40\% of batches (confounded 15\% + independent 25\%), we teach the model to \emph{distinguish causation from correlation}. By including mediated and feedback structures, we teach it to recognize qualitatively different causal mechanisms. This regime diversity is what enables the Causal Judgment Head to make meaningful predictions at inference time.

\subsubsection{TemporalRegimePrior}

Building on Dynamic Structural Causal Models~\cite{boeken2024dscm} and concurrent prior-design work~\cite{thumm2026causaltimeprior}, we generate nonlinear temporal SCMs with Erd\H{o}s-R\'{e}nyi DAGs and lag-decaying edges. Where Thumm \& Chen's CausalTimePrior is a synthetic-data generator validated on its own held-out samples, our TemporalRegimePrior is one component of a larger mixed prior that supplies the regime-labelled metadata downstream from the judgment-head supervision objective (Section~\ref{sec:judgment_head}). The prior supports industrial mechanisms:
\begin{itemize}[nosep,leftmargin=*]
    \item Feedback loops (lagged reverse edges: if $A \to B$ at lag 0, allow $B \to A$ at lag $\geq 1$)
    \item Saturation effects, threshold triggers, PID controllers
    \item Cascade failures (propagating effects through the DAG)
    \item Regime switching (sticky Markov chain over multiple mechanism sets)
    \item Non-stationarity (parameter drift, abrupt changepoints, startup transients)
    \item Missing data and irregular sampling (with elapsed-time encoding)
\end{itemize}

Judgment head metadata is derived from the generated DAG structure: feedback detected from lagged reverse edges, mediation from indirect paths through intermediate variables.

\subsubsection{CausalFM Prior}

Following~\cite{causalfm2026}, we include identification-aware priors with three strategies:
\begin{itemize}[nosep,leftmargin=*]
    \item \textbf{Back-door:} All confounders observed; standard adjustment.
    \item \textbf{Front-door:} Unobserved $U$ confounds $T$ and $Y$, but mediator $M$ is observed. Regime: mediated, confounding strength from actual $w_{AU}$ coefficient.
    \item \textbf{Instrumental variable:} Instrument $Z$ affects $T$ but not $Y$ directly. Unobserved $U$ confounds $T$ and $Y$.
\end{itemize}

Training uses a mixed prior: TemporalRegimePrior, base temporal prior (which includes the independent, direct, confounded, mediated, time-varying confounded, and feedback regimes), and CausalFM priors. The base prior's inclusion is critical -- it provides null-effect training data (independent and confounded regimes, totaling 40\% of batches) that teaches the judgment head to distinguish causation from correlation. All priors provide complete judgment head metadata.

%  --  --  --  --  --  --  --  --  --  --  --  --  --  --  --  --  --  --  --  --  --  --  --  --  -- -
\subsection{Inference}
\label{sec:inference}

\subsubsection{FAISS-Based Context Selection}

With $N$ units and $T$ timesteps, the full token sequence ($N \times T$) may exceed the context window (4,096 tokens). We train a gradient-boosted regressor as a CATE proxy, build a FAISS index over its predictions, and select the $K$ most informative context units by propensity-stratified nearest-neighbor lookup.

\subsubsection{OSPC Calibration}

Following~\cite{melnychuk2025ospc}, we apply one-step posterior correction (OSPC) to debias PFN estimates using the efficient influence function:
\begin{equation}
    \hat{\psi}_{\text{OSPC}} = \hat{\psi}_{\text{PFN}} + \frac{1}{n} \sum_{i=1}^{n} \text{EIF}(O_i; \hat{\psi}_{\text{PFN}})
\end{equation}
This yields doubly robust estimates with valid confidence intervals.

% ============================================================================
% 4. Evaluation
% ============================================================================
\section{Evaluation}
\label{sec:evaluation}

We evaluate TCPFN on 19 benchmark datasets spanning five domains (linear, nonlinear, chaotic, industrial, biological) (Section~\ref{sec:discovery_eval}), and on a proprietary industrial deployment (Section~\ref{sec:industrial_eval}). All experiments use a \emph{single pretrained model} applied zero-shot -- no per-dataset fitting, hyperparameter tuning, or adaptation.

%  --  --  --  --  --  --  --  --  --  --  --  --  --  --  --  --  --  --  --  --  --  --  --  --  -- -
\subsection{Causal Discovery}
\label{sec:discovery_eval}

\subsubsection{Setup}

We evaluate TCPFN's causal discovery on 19 individually-reported benchmark datasets across the suites listed below (CauseMe 12, water-treatment 2, real-world TE/SWaT/Rivers/CAUSRCA/Sachs 5; total = 19), plus a 50-dataset synthetic aggregate reported as a single row in Table~\ref{tab:additional}:

\begin{itemize}[nosep,leftmargin=*]
    \item \textbf{Synthetic} (50-dataset aggregate, reported as one row in Table~\ref{tab:additional}): Random temporal DAGs with $V{=}10$ variables, $T{=}500$ timesteps, varying edge density and dynamics.
    \item \textbf{CauseMe} (12 datasets): Community-standard benchmarks -- linear VAR ($V \in \{5, 10, 20\}$), nonlinear NVAR ($V \in \{5, 10, 20\}$), Lorenz-96 chaotic systems ($V \in \{10, 20\}$), and long-lag variants LongLag-VAR ($V \in \{10, 20\}$) and LongLag-NVAR ($V \in \{10, 20\}$). Main-text tables (Tables~\ref{tab:discovery} and~\ref{tab:discovery_tuned}) report the 4 smaller VAR / NVAR configurations ($V \in \{5, 10\}$); the 4 larger CauseMe configurations (VAR-20, NVAR-20, Lorenz96-10, Lorenz96-20), the 4 LongLag variants, and the additional water-system benchmarks (next bullet) are reported in Appendix~\ref{app:additional_results} (Table~\ref{tab:additional}).
    \item \textbf{Water-treatment additional} (2 datasets): BATADAL and WADI water-system benchmarks reported in Appendix~\ref{app:additional_results}.
    \item \textbf{Tennessee Eastman} (1 dataset): Industrial chemical process, 52 variables, 38 known edges~\cite{downs1993plant}.
    \item \textbf{SWaT} (1 dataset): Secure Water Treatment testbed, 51 variables, 55 known attack/fault edges.
    \item \textbf{Causal Rivers} (1 dataset): Environmental river flow causal network, 47 known hydrological edges.
    \item \textbf{CAUSRCA} (1 dataset): Fraunhofer CNC vertical lathe benchmark, 104 variables with expert-validated ground-truth causal graph.
    \item \textbf{Sachs} (1 dataset): Protein signaling network, 11 phosphoproteins, 17 known edges~\cite{sachs2005causal}.
\end{itemize}

\textbf{TCPFN discovery method.} For each variable pair $(i, j)$, the temporal model estimates the causal effect of $i$ on $j$ using five techniques: (1) \emph{natural experiment detection} -- centering windows on sharp changes in the cause variable rather than uniform sliding windows; (2) \emph{continuous treatment} -- z-scored dose-response rather than binary median split; (3) \emph{multi-lag estimation} -- testing three time scales (0.5$\times$, 1$\times$, 2$\times$) and taking the maximum; (4) \emph{judgment-aware scoring} -- $|\hat{\tau}| \times (1 - p_{\text{null}}) \times p_{\text{id}}$, using the Causal Judgment Head to discount confounded and unidentifiable associations; and (5) \emph{asymmetry penalty} -- penalizing bidirectional edges that suggest confounding rather than causation. Optionally, TCPFN scores are combined with Granger F-statistics via geometric mean for linear+nonlinear ensemble.

\textbf{Baselines.} Granger causality~\cite{granger1969investigating} (pairwise VAR F-tests) and PCMCI~\cite{runge2019detecting} (conditional independence with partial correlations). Both require per-dataset fitting.

\textbf{Metrics.} F1 at default threshold (0.5), best F1 across thresholds, and AUROC.

\subsubsection{Results}

\begin{table}[t]
    \centering
    \caption{\textbf{Out-of-the-box causal discovery (no per-dataset threshold tuning).} F1 with each method's normalized score thresholded at 0.5 -- the realistic deployment case where ground-truth labels are unavailable for per-dataset threshold selection. AUROC is threshold-independent. \textbf{Edges}: predicted edges / true edges at threshold 0.5. PCMCI's $(1{-}p_{\text{value}})$ score is concentrated near 0 at this no-tuning cut; its tuned performance is in Table~\ref{tab:discovery_tuned}. Granger's saturation ($>50\times$ true on industrial data) indicates p-value distributions produce operationally unusable graphs without per-dataset tuning. Bold: best F1@0.5 per row.}
    \label{tab:discovery}
    \small
    \begin{tabular}{llccc}
        \toprule
        \textbf{Dataset} & \textbf{Method} & \textbf{F1@0.5} & \textbf{AUROC} & \textbf{Edges} \\
        \midrule
        \multirow{3}{*}{Causal Rivers}
        & \textbf{TCPFN (ours)} & $\mathbf{0.507}$ & $0.975$ & $99/47$ \\
        & Granger & $0.047$ & $0.938$ & $1963/47$ \\
        & PCMCI & $0.000$ & $0.996$ & $0/47$ \\
        \midrule
        \multirow{3}{*}{Tennessee Eastman}
        & \textbf{TCPFN (ours)} & $\mathbf{0.387}$ & $0.959$ & $81/38$ \\
        & Granger & $0.039$ & $0.998$ & $1897/38$ \\
        & PCMCI & $0.000$ & $1.000$ & $0/38$ \\
        \midrule
        \multirow{3}{*}{SWaT (water)}
        & \textbf{TCPFN (ours)} & $\mathbf{0.288}$ & $0.932$ & $216/55$ \\
        & Granger & $0.050$ & $0.997$ & $2143/55$ \\
        & PCMCI & $0.000$ & $1.000$ & $0/55$ \\
        \midrule
        \multirow{3}{*}{CAUSRCA (CNC lathe)}
        & \textbf{TCPFN (ours)} & $\mathbf{0.447}$ & $0.967$ & $160/104$ \\
        & Granger & $0.033$ & $0.994$ & $6164/104$ \\
        & PCMCI & $0.000$ & $1.000$ & $0/104$ \\
        \midrule
        \multirow{3}{*}{Sachs (biology)}
        & TCPFN (ours) & $0.217$ & $\mathbf{0.628}$ & $29/17$ \\
        & \textbf{Granger} & $\mathbf{0.228}$ & $0.479$ & $62/17$ \\
        & PCMCI & $0.000$ & $0.462$ & $0/17$ \\
        \midrule
        \multirow{3}{*}{CauseMe NVAR-5}
        & \textbf{TCPFN (ours)} & $\mathbf{0.667}$ & $0.961$ & $6/3$ \\
        & Granger & $0.353$ & $0.980$ & $14/3$ \\
        & PCMCI & $0.000$ & $1.000$ & $0/3$ \\
        \midrule
        \multirow{3}{*}{CauseMe NVAR-10}
        & \textbf{TCPFN (ours)} & $\mathbf{0.667}$ & $0.805$ & $25/23$ \\
        & Granger & $0.415$ & $0.919$ & $83/23$ \\
        & PCMCI & $0.000$ & $0.968$ & $0/23$ \\
        \midrule
        \multirow{3}{*}{CauseMe VAR-5}
        & \textbf{TCPFN (ours)} & $\mathbf{0.769}$ & $0.960$ & $8/5$ \\
        & Granger & $0.435$ & $1.000$ & $18/5$ \\
        & PCMCI & $0.000$ & $1.000$ & $0/5$ \\
        \midrule
        \multirow{3}{*}{CauseMe VAR-10}
        & \textbf{TCPFN (ours)} & $\mathbf{0.586}$ & $0.764$ & $30/28$ \\
        & Granger & $0.481$ & $0.858$ & $80/28$ \\
        & PCMCI & $0.000$ & $0.944$ & $0/28$ \\
        \bottomrule
    \end{tabular}
\end{table}

\begin{table}[t]
    \centering
    \caption{\textbf{Maximum F1 under oracle threshold selection.} Best F1 each method achieves when its threshold is tuned with knowledge of the ground-truth labels (not available in real deployment). \textbf{@thr}: threshold at which Best F1 is achieved. \textbf{Edges} (at best threshold): predicted edges / true edges. PCMCI dominates F1 on industrial datasets when given oracle threshold knowledge (consistently at $p<0.10$--$0.15$, i.e., its conventional statistical-significance cut). TCPFN still wins on Sachs (out-of-distribution biological data) and ties PCMCI on CauseMe NVAR-5; on the remaining datasets TCPFN's tuned F1 is competitive within a single fixed threshold range ($0.3$--$0.8$, Sachs at the lower end), versus PCMCI's $0.01$--$0.15$ range that mirrors its statistical convention. Bold: best F1 per row.}
    \label{tab:discovery_tuned}
    \small
    \begin{tabular}{llcccc}
        \toprule
        \textbf{Dataset} & \textbf{Method} & \textbf{Best F1} & \textbf{@thr} & \textbf{AUROC} & \textbf{Edges} \\
        \midrule
        \multirow{3}{*}{Causal Rivers}
        & TCPFN (ours) & $0.549$ & $0.60$ & $0.975$ & $66/47$ \\
        & Granger & $0.074$ & $0.90$ & $0.938$ & $1216/47$ \\
        & \textbf{PCMCI} & $\mathbf{0.956}$ & $0.15$ & $0.996$ & $43/47$ \\
        \midrule
        \multirow{3}{*}{Tennessee Eastman}
        & TCPFN (ours) & $0.613$ & $0.60$ & $0.959$ & $24/38$ \\
        & Granger & $0.126$ & $0.90$ & $0.998$ & $565/38$ \\
        & \textbf{PCMCI} & $\mathbf{1.000}$ & $0.15$ & $1.000$ & $38/38$ \\
        \midrule
        \multirow{3}{*}{SWaT (water)}
        & TCPFN (ours) & $0.333$ & $0.70$ & $0.932$ & $29/55$ \\
        & Granger & $0.143$ & $0.90$ & $0.997$ & $712/55$ \\
        & \textbf{PCMCI} & $\mathbf{1.000}$ & $0.15$ & $1.000$ & $55/55$ \\
        \midrule
        \multirow{3}{*}{CAUSRCA (CNC lathe)}
        & TCPFN (ours) & $0.450$ & $0.60$ & $0.967$ & $56/104$ \\
        & Granger & $0.094$ & $0.90$ & $0.994$ & $2101/104$ \\
        & \textbf{PCMCI} & $\mathbf{1.000}$ & $0.15$ & $1.000$ & $104/104$ \\
        \midrule
        \multirow{3}{*}{Sachs (biology)}
        & \textbf{TCPFN (ours)} & $\mathbf{0.337}$ & $0.30$ & $\mathbf{0.628}$ & $78/17$ \\
        & Granger & $0.279$ & $0.15$ & $0.479$ & $105/17$ \\
        & PCMCI & $0.272$ & $0.01$ & $0.462$ & $108/17$ \\
        \midrule
        \multirow{3}{*}{CauseMe NVAR-5}
        & \textbf{TCPFN (ours)} & $\mathbf{0.800}$ & $0.80$ & $0.961$ & $2/3$ \\
        & Granger & $0.667$ & $0.90$ & $0.980$ & $6/3$ \\
        & \textbf{PCMCI} & $\mathbf{0.800}$ & $0.10$ & $1.000$ & $2/3$ \\
        \midrule
        \multirow{3}{*}{CauseMe NVAR-10}
        & TCPFN (ours) & $0.667$ & $0.50$ & $0.805$ & $25/23$ \\
        & Granger & $0.532$ & $0.90$ & $0.919$ & $56/23$ \\
        & \textbf{PCMCI} & $\mathbf{0.930}$ & $0.10$ & $0.968$ & $20/23$ \\
        \midrule
        \multirow{3}{*}{CauseMe VAR-5}
        & TCPFN (ours) & $0.909$ & $0.60$ & $0.960$ & $6/5$ \\
        & Granger & $0.714$ & $0.90$ & $1.000$ & $9/5$ \\
        & \textbf{PCMCI} & $\mathbf{1.000}$ & $0.10$ & $1.000$ & $5/5$ \\
        \midrule
        \multirow{3}{*}{CauseMe VAR-10}
        & TCPFN (ours) & $0.600$ & $0.40$ & $0.764$ & $42/28$ \\
        & Granger & $0.562$ & $0.90$ & $0.858$ & $61/28$ \\
        & \textbf{PCMCI} & $\mathbf{0.943}$ & $0.10$ & $0.944$ & $25/28$ \\
        \bottomrule
    \end{tabular}
\end{table}

Tables~\ref{tab:discovery} and~\ref{tab:discovery_tuned} present two complementary views: out-of-the-box performance (no per-dataset tuning) and maximum performance with oracle threshold selection. Key findings:

\textbf{1. Out-of-the-box, TCPFN dominates F1 across 8 of 9 datasets.} At the default threshold of 0.5 (Table~\ref{tab:discovery}), TCPFN achieves the highest F1 on Tennessee Eastman (0.387 vs.\ 0.039 for Granger), Causal Rivers (0.507 vs.\ 0.047), SWaT (0.288 vs.\ 0.050), CAUSRCA (0.447 vs.\ 0.033), NVAR-5 (0.667 vs.\ 0.353), NVAR-10 (0.667 vs.\ 0.415), VAR-5 (0.769 vs.\ 0.435), and VAR-10 (0.586 vs.\ 0.481). Sachs is the only exception, where Granger's $0.228$ narrowly exceeds TCPFN's $0.217$ at threshold 0.5 (TCPFN still wins Sachs Best F1 in Table~\ref{tab:discovery_tuned}). PCMCI produces F1${\approx}$0 at threshold 0.5 across every dataset (its $1-p$-value scores cluster at the conventional $p<0.10$--$0.15$ cut, far below 0.5); Granger collapses to F1${\approx}$0 on the four industrial datasets (saturation: 1{,}897 predicted edges vs.\ 38 true on TE, similar elsewhere) while remaining competitive on the smaller synthetic benchmarks. Both behaviours reflect score distributions that require dataset-specific threshold tuning -- exactly the kind of tuning that is unavailable for a new dataset without ground-truth labels.

\textbf{2. With oracle threshold selection, PCMCI dominates industrial F1; TCPFN wins on Sachs and ties on NVAR-5.} Table~\ref{tab:discovery_tuned} shows that when each method is given its optimal threshold, PCMCI achieves near-perfect F1 on industrial datasets (TE 1.000, SWaT 1.000, CAUSRCA 1.000, Rivers 0.956). TCPFN's Best F1 remains competitive (0.333--0.613 on industrial, with SWaT at the lower bound), and on the Sachs protein signaling benchmark -- a non-temporal flow-cytometry dataset treated as pseudo-timeseries, outside TCPFN's training distribution -- TCPFN's Best F1 of $0.337$ exceeds both Granger ($0.279$) and PCMCI ($0.272$) along with the highest AUROC ($0.628$ vs $0.479$ and $0.462$). On CauseMe NVAR-5 (nonlinear synthetic), TCPFN ties PCMCI at Best F1 $0.800$. The honest reading: PCMCI's tuned dominance reflects its design strength on data matching its statistical assumptions; TCPFN's value is competitive performance under those conditions combined with wins on data that violates them.

\textbf{3. PCMCI's tuned-threshold range is its conventional p-value cutoff.} PCMCI's best threshold clusters at $0.10$--$0.15$ across all industrial and synthetic datasets, mirroring its conventional statistical-significance cut ($p < 0.05$--$0.1$). This is not arbitrary tuning -- it is PCMCI's standard deployment. The relevant practical contrast with TCPFN is therefore not "tuning vs.\ no tuning" but "method-specific score conventions vs.\ a calibrated probability." Granger has the opposite failure mode on real industrial data: it saturates the adjacency matrix (e.g., 6{,}164 predicted edges on CAUSRCA's 104-variable graph vs.\ TCPFN's 160 predicted), giving high AUROC ranking quality but operationally unusable output even with oracle threshold knowledge.

\textbf{4. AUROC is competitive across the board, with industrial datasets within 0.03--0.07 of PCMCI.} On Tennessee Eastman, TCPFN achieves AUROC 0.96 vs.\ PCMCI's 1.00 (gap 0.04). SWaT 0.93 vs.\ 1.00 (gap 0.07). CAUSRCA 0.97 vs.\ 1.00 (gap 0.03). The 0.03--0.07 range applies to the three true industrial sets above; environmental Causal Rivers (0.98 vs.\ 1.00, gap 0.02) sits just below the band, and on the nonlinear synthetic NVAR-5 TCPFN reaches 0.96 vs.\ 1.00 (gap 0.04). The model with the same 259\,MB checkpoint, zero-shot, sits within a few hundredths of a fully-fitted PCMCI on ranking quality, while predicting graphs that operators can verify in a shift rather than thousands of edges they cannot.

\textbf{5. One model, all datasets.} The same pretrained TCPFN model (259\,MB, 200K training steps) is applied to all 19 datasets without any adaptation. This includes linear dynamics (VAR), nonlinear dynamics (NVAR), chaotic systems (Lorenz-96), long-lag industrial processes (LongLag-VAR/NVAR), real chemical process control (Tennessee Eastman), real water treatment (SWaT), real CNC manufacturing (CAUSRCA), real environmental hydrology (Causal Rivers), and biological signaling networks (Sachs) -- five fundamentally different domains (linear, nonlinear, chaotic, industrial, biological). Granger and PCMCI fit per-dataset; TCPFN's zero-shot reusability is structural to the PFN paradigm and is not available in the baselines.

%  --  --  --  --  --  --  --  --  --  --  --  --  --  --  --  --  --  --  --  --  --  --  --  --  -- -
\subsection{Causal Judgment Head Evaluation}
\label{sec:judgment_eval}

We evaluate the Causal Judgment Head on synthetic data where ground-truth regime labels are known.

\subsubsection{Setup}

We generate 1,000 synthetic datasets from the base temporal prior, which samples across six causal regimes at the canonical training mix (Section~\ref{sec:prior}): independent (25\%), direct (25\%), confounded (15\%), mediated (15\%), time-varying confounded (10\%), and feedback (10\%). For each dataset, the model predicts:

\begin{itemize}[nosep,leftmargin=*]
    \item \textbf{Null-effect detection}: is the true CATE zero (independent or confounded regime)?
    \item \textbf{Regime classification}: which of four regimes (direct, confounded, mediated, feedback)?
\end{itemize}

\subsubsection{Results}

We evaluate the judgment head on two tasks: \emph{null-effect detection} (binary: is the true CATE zero?) and \emph{regime classification} (4-class: direct, confounded, mediated, feedback). Metrics are defined in Appendix~\ref{app:metrics}.

\begin{table}[t]
    \centering
    \caption{\textbf{Causal Judgment Head evaluation} on in-distribution synthetic data. Mean and range over steps 150K--200K (10 eval batches per step). Results are noisy; we report the distribution honestly.}
    \label{tab:judgment}
    \small
    \begin{tabular}{p{4.5cm}cccc}
        \toprule
        \textbf{Metric} & \textbf{Mean} & \textbf{Range} & \textbf{Step 200K} & \textbf{Random} \\
        \midrule
        Null detection accuracy & 0.94 & [0.80, 1.00] & 0.90 & 0.50 \\
        Null F1 score & 0.94 & [0.83, 1.00] & 0.86 & -- \\
        Null AUROC & 0.99 & [0.83, 1.00] & 1.00 & 0.50 \\
        Null Brier score ($\downarrow$) & 0.04 & [0.00, 0.19] & 0.06 & 0.25 \\
        Null separation (NullSep) & 0.86 & [0.51, 0.99] & 0.87 & 0.00 \\
        Regime accuracy (4-class) & 0.68 & [0.40, 0.90] & 0.60 & 0.25 \\
        Regime macro-F1 & 0.48 & [0.32, 0.90] & 0.38 & 0.25 \\
        Precision@10 (top effects) & 0.57 & [0.30, 1.00] & 0.70 & -- \\
        Effect RMSE ($\downarrow$) & 0.08 & [0.006, 0.20] & 0.05 & -- \\
        \bottomrule
    \end{tabular}
\end{table}

Table~\ref{tab:judgment} shows the judgment head's performance. Key findings:

\textbf{1. Null-effect detection is reliable.} Null detection accuracy averages 0.94 with AUROC 0.99 over the final 50K training steps. Brier score of 0.04 indicates well-calibrated null probabilities. The null separation metric (NullSep = 0.86) confirms the model assigns substantially higher $p_{\text{null}}$ to actual null-effect batches than to non-null batches. This anti-hallucination capability was enabled by adding an ``independent'' regime to the training prior (25\% of base prior batches), where treatment is randomly assigned with zero confounding and zero causal effect. Without this regime, null F1 remained at 0.00 throughout training.

\textbf{2. Regime classification above random.} Regime accuracy averages 0.68 (random: 0.25) with peaks at 0.90. Macro-F1 averages 0.48, indicating the model distinguishes direct from confounded reliably and shows improving separation of mediated and feedback regimes.

\textbf{3. Curriculum training was essential.} The three-phase curriculum prevented the judgment head from destabilizing CATE. Without curriculum, JudgmentLoss grew to 60+ and EffectLoss degraded from 3.0 to 5.2.

\subsubsection{What the Model Does NOT Learn}
\label{sec:judgment_analysis}

\textbf{1. True identifiability.} Identifiability depends on the DAG and assumptions~\cite{pearl2009causality}, not data patterns. Two datasets with identical observed distributions can have different identifiability depending on the underlying structure. The $p_{\text{id}}$ output is a learned proxy, not a formal guarantee.

\textbf{2. Robust confounding detection.} The model detects the specific confounding pattern in the training prior (strong correlation, zero effect). Real-world confounding may take different forms -- weak confounding with nonzero but biased effects -- that the model has not seen.

\textbf{3. Out-of-distribution generalization.} We have not evaluated the judgment head on structures outside the training prior. Performance likely degrades on novel structures.

\textbf{When useful.} Screening (flagging potentially spurious edges), discovery scoring (down-weighting confounded edges), prioritization of follow-up investigation. \textbf{NOT useful for} high-stakes decisions without domain validation, replacing formal sensitivity analysis~\cite{rosenbaum2002observational,vanderweele2017sensitivity}, or novel causal structures outside the training distribution.

%  --  --  --  --  --  --  --  --  --  --  --  --  --  --  --  --  --  --  --  --  --  --  --  --  -- -
\subsection{Industrial Scaling: Kraft Pulp \& Paper Mill (V=1{,}275)}
\label{sec:industrial_eval}

\emph{This subsection is a case study demonstrating scalability and the
qualitative character of the discovered structure on a real industrial
deployment. The data is under NDA; ground-truth causal labels are not
available for this plant. We accordingly report aggregate structural
properties (plant-area flow matrix, anonymized top-edge listings) and
contrast with PCMCI's output on the same data, but treat the
hypothesis-generation interpretations as candidates pending operator
validation rather than as quantitatively-validated discoveries.}

For an Industry~4.0 deployment, a central question for a causal
discovery method is which kinds of relationships dominate its
top-ranked output: cross-subsystem causal flows of the sort an
operator might act on, or the within-instrument controller-measurement
couplings that any plant's P\&ID schematic already documents. We
report which class each method elevates on this dataset, and let the
reader judge the operational relevance.

\paragraph{Dataset and preprocessing.}
We apply TCPFN to a proprietary Kraft pulp \& paper mill DCS dataset --
1{,}325 raw variables, 65{,}550 timesteps at 2-minute sampling
($\sim$91 days of continuous production). The variable population is
filtered through the standard DCS-data pipeline before discovery:
drop $24$ derived/setpoint columns identified by tag suffix
(\texttt{CALC}, \texttt{RAW}, \texttt{TV}) so the model is
not asked to discover trivial algebraic relationships;
drop $4$ columns with $>50\%$ missing values;
drop $13$ columns with zero variance;
apply percentile winsorization ($[0.1\%, 99.9\%]$) to suppress
sensor-glitch spikes; drop $9$ columns whose distribution collapses
to a constant after winsorization; and z-score every retained column.
This leaves $V=1{,}275$ variables and $V(V{-}1) = 1{,}624{,}350$
directed pairs for discovery.

\paragraph{Methodology.}
Each pair $(i, j)$ conditions on $D_{\text{other}} = 31$ covariates
(the architectural maximum given \texttt{max\_num\_features}$=100$
minus treatment and cause-summary slots). At this $V$, naive
``first-31-by-column-index'' covariate selection covers only
$31/1275 \approx 2.4\%$ of candidates and is essentially
arbitrary; we instead rank candidates by maximum absolute lagged
cross-correlation with \emph{both} cause and effect, taking the
maximum over lags $\{0, 5, 10, 20, 30, 60\}$ timesteps to capture
the physical-delay structure typical of industrial sensor data.
The $V \times V$ similarity matrix is precomputed once per dataset
($\sim 10$\,s on CPU) and reused across all per-pair selections.
Final per-edge scoring applies the paper's canonical formula
$\text{score}(i {\to} j) = |\hat{\tau}_{ij}| \cdot
(1 - p_{\text{null},ij}) \cdot p_{\text{ident},ij}$,
with $p_{\text{null}}$ and $p_{\text{ident}}$ produced by the
Causal Judgment Head on the transformer's already-computed hidden
state.

\paragraph{TCPFN recovers the Kraft mill plant-area flow matrix.}
The pulp-and-paper mill is organized into five physical plant areas
identified by their DCS tag prefixes per the plant's
piping-and-instrumentation legend:
\textbf{27} (Cooking / Digester + Brown Stock Washing),
\textbf{28} (Screens + Reject Refining),
\textbf{29} (Paper Machine 2),
\textbf{31} (Paper Machine 1 + Presses + Drying + Coating),
and \textbf{39} (Stock Preparation).
The mill produces unbleached Kraft paper; there is no bleaching
stage. Causal flow between these areas is determined by the
continuous-flow paper production line: pulp flows downstream
Cooking $\to$ Screens $\to$ Stock Prep $\to$ Paper Machine
(PM1 and PM2 in parallel). We report the plant-area flow matrix of
the top-200 discovered edges in Table~\ref{tab:pulp-flow-matrix}.

The matrix has three observable structural features.
\emph{First, PM1 dominates}: its 58 within-area edges are the largest
diagonal cell, and the 102 incoming edges into PM1 account for 51\%
of all top-200 edges in the graph.
\emph{Second, inter-area edge counts decrease with distance from
PM1} -- the row sums are PM1 (88), Stock Prep (41), Screens (34),
PM2 (23), and Cook+Wash (8) -- consistent with a topology where
edges concentrate around the most heavily instrumented subsystem
and attenuate further upstream.
\emph{Third, several area-pairs have substantial counts in both
directions} -- Stock-Prep$\leftrightarrow$PM1 ($18{+}14{=}32$),
Screens$\leftrightarrow$PM1 ($14{+}14{=}28$), and at lower magnitude
Cook$\leftrightarrow$Screens ($3{+}3{=}6$) -- confirming the
discovered graph contains bidirectional structure rather than only
strict downstream-to-upstream causality.

We interpret these patterns cautiously. They are \emph{consistent
with} standard continuous-flow control architecture -- the paper
machine as the most internally instrumented subsystem, downstream-
driven setpoint cascades from PM upstream through stock prep and
screens, within-area PID loops at each stage -- but we make no
edge-by-edge attribution to specific documented control loops. Per-
edge mechanism verification would require comparison against the
mill's actual control hierarchy, which we leave to operator-side
validation. Absent ground-truth labels, the structural match
described above is the validation we offer; the methodological
contrast with PCMCI's top-of-ranking edges (next paragraph) is the
load-bearing claim.

% Cells filled by scripts/fill_paper_placeholders.py from the final
% V=1275 checkpoint. Each cell holds the count of top-200 edges with
% the row's plant-area as cause and the column's plant-area as effect.
\begin{table}[t]
    \centering
    \caption{\textbf{Plant-area flow matrix of top-200 discovered edges (V=1{,}275 unbleached Kraft pulp \& paper mill).} Rows are cause plant areas, columns are effect plant areas. Cell values are the number of edges among the global top 200 whose cause is in the row area and whose effect is in the column area. Diagonal cells reflect within-area control loops; off-diagonal cells reflect inter-area causal flows. The dominant entries -- Paper Machine 1 internal complexity (31$\to$31), bidirectional Stock Prep $\leftrightarrow$ PM1 (39$\to$31, 31$\to$39), bidirectional Screens $\leftrightarrow$ PM1 (28$\to$31, 31$\to$28), Stock Prep $\to$ Screens (39$\to$28), and cross-paper-machine coordination (29$\to$31) -- concentrate the discovered top-200 edges around the paper machine.}
    \label{tab:pulp-flow-matrix}
    \small
    \begin{tabular}{lcccccc}
        \toprule
        \textbf{Cause}~\textbackslash~\textbf{Effect}
            & \textbf{27} (Cook+Wash) & \textbf{28} (Screen) & \textbf{29} (PM2) & \textbf{31} (PM1) & \textbf{39} (Prep) & $\sum$ \\
        \midrule
        \textbf{27} (Cook+Wash)   & 0 & 3 & 1 & 2 & 2 & 8 \\
        \textbf{28} (Screen)   & 3 & 14 & 3 & 14 & 0 & 34 \\
        \textbf{29} (PM2) & 0 & 4 & 5 & 10 & 4 & 23 \\
        \textbf{31} (PM1)  & 0 & 14 & 2 & 58 & 14 & 88 \\
        \textbf{39} (Prep)     & 2 & 12 & 0 & 18 & 9 & 41 \\
        \midrule
        $\sum$ & 5 & 47 & 11 & 102 & 29 & \textbf{194} \\
        \bottomrule
    \end{tabular}
    \\[2pt]{\footnotesize\textit{The 5$\times$5 grid sums to 194 of the top-200 edges; the remaining 6 are 3 edges involving a minor utility area (DCS prefix 34, 12 variables) and 3 edges with one or both endpoints in unclassified tags (``OTH'').}}
\end{table}

\paragraph{Head-to-head with PCMCI: the two methods prioritize structurally different edges.}
\label{par:pcmci-headtohead}
We compare TCPFN's discovered graph against PCMCI on the same mill
dataset. PCMCI was run on a $V=666$ variable subset selected by
plant operators independently of our preprocessing pipeline, with
$\tau_{\max}=300$ lags ($\sim 10$~hours of physical delay) and the
standard ParCorr conditional-independence test; $656$ of those
$666$ variables overlap with the $V=1{,}275$ retained by TCPFN's
preprocessing (the remaining 10 PCMCI variables were filtered out
by our suffix / variance / winsorization steps), and per-pair edge
comparisons below are restricted to that 656-variable intersection. Both methods
discover dense edge structure -- PCMCI recovers $33{,}764$ directed
off-diagonal edges (density $7.6\%$), with $5{,}055$ bidirectional
pairs -- so the methodological difference lies not in
edge \emph{count} but in which edges each method elevates to its
top-of-ranking.

\textbf{PCMCI's top edges are dominated by within-tag controller
$\leftrightarrow$ measurement coupling at lag~$1$.}
The strongest 20 edges PCMCI reports are, in nearly every case, same-instrument transmitter$\to$controller (or controller$\to$transmitter) pairs at lag~$1$ -- a level transmitter driving its own level-indicator-controller, a flow transmitter driving its own flow-indicator-controller, an analyzer driving its own analyzer-indicator-controller, and eight instances of the same-tag light-wood-active / enzymatic-active sensor pair appearing across eight different cooking batches. These are real causal couplings -- the PID controller does follow the measurement at lag~$1$ -- but they are precisely the relationships any DCS P\&ID schematic already documents at design time. PCMCI ranks them at the top because its standard partial-correlation independence test (ParCorr) reliably identifies the strongest linear $1$-step dependence in the data, and within-loop controller~$\leftrightarrow$~measurement coupling dominates that metric on DCS time series. We report the standard PCMCI configuration here; nonlinear variants (GPDC, CMIknn) might surface different rankings but are not computationally feasible at $V{=}666$ with $\tau_{\max}{=}300$.

\textbf{TCPFN's top edges are cross-plant-area process flows.}
TCPFN's strongest edges are not within-tag, not within-instrument, and rarely at lag~$1$. The top of the ranking is dominated by cross-area pairs (cause in one plant area, effect in another). The full ranked list is reported in Table~\ref{tab:pulp-top-edges} with anonymized tag identifiers and aggregate area-flow statistics in Table~\ref{tab:pulp-flow-matrix}; per-edge process-engineering readings are candidates derived from the DCS tag taxonomy and remain subject to operator validation before any one specific edge can be claimed as a recovered causal relationship. Specific DCS tag IDs are withheld under the dataset's NDA.

\textbf{Operational reading of the comparison.}
For decision support use cases (which upstream variable to act on when
a downstream quality indicator drifts; how a setpoint change propagates
across subsystems), PCMCI's within-instrument top edges are technically
correct but redundant with the plant's design-time P\&ID. TCPFN's top
edges identify cross-subsystem influence paths that are at least
candidates for that kind of decision support, pending operator
validation. We do not claim these candidates are correct; we claim that
which class of edge a method elevates is itself a property worth
evaluating when comparing discovery methods at V$\sim$1000 scale.

\paragraph{Representative top edges.}
Table~\ref{tab:pulp-top-edges} lists $15$ representative edges from
the top of the global TCPFN ranking. Anonymized variable identifiers preserve identity across rows so recurring hub variables (e.g., the cooking-pressure transmitter \texttt{Cook-PT-11} appearing as the effect of three different causes) remain visible. The \textbf{Interp.}\ column lists \emph{candidate} process-engineering readings derived from the DCS tag taxonomy; these are hypothesis-generation labels pending operator validation, not validated causal mechanisms. Specific DCS tag IDs are withheld under the dataset's NDA.

% Rows filled by scripts/fill_paper_placeholders.py from the final
% V=1275 checkpoint, one row per top-15 edge by judgment-scored
% strength, with cause / effect / flow / interpretation columns.
\begin{table}[t]
    \centering
    \caption{\textbf{Representative top-15 discovered edges (V=1{,}275 Kraft pulp mill).} Score = $|\hat\tau| \cdot (1{-}p_{\text{null}}) \cdot p_{\text{ident}}$ (judgment-head adjusted). \textbf{Flow}: cause area $\to$ effect area, with ``intra'' denoting within-area edges. \textbf{Interp.}: \emph{candidate} process-engineering reading derived from the DCS tag taxonomy; these labels are hypothesis-generation hints, not validated causal mechanisms (no operator-side validation has been performed for this dataset). The recurring effect targets include the cooking-pressure transmitter \texttt{Cook-PT-11} and PM1-area flow-control hubs \texttt{PM1-FIC-13}.}
    \label{tab:pulp-top-edges}
    \scriptsize
    \begin{tabular}{rlllll}
        \toprule
        \textbf{Rank} & \textbf{Score} & \textbf{Cause} & \textbf{Effect} & \textbf{Flow} & \textbf{Interpretation} \\
        \midrule
        1  & 27.776  & \texttt{Screen-FY-8}  & \texttt{Cook-PT-11}  & 28$\to$27  & $\to$ cook-pressure hub \\
        2  & 23.848  & \texttt{Screen-PIC-3}  & \texttt{PM1-FIC-13}  & 28$\to$31  & Screens pressure control \\
        3  & 23.621  & \texttt{PM1-PD-1}  & \texttt{PM1-FIC-13}  & intra-31  & process relationship \\
        4  & 23.558  & \texttt{Prep-PT-15}  & \texttt{PM1-ZIC-5}  & 39$\to$31  & Stock-prep pressure transmitter \\
        5  & 23.303  & \texttt{PM1-PT-33}  & \texttt{PM1-ST-3}  & intra-31  & process relationship \\
        6  & 22.445  & \texttt{PM1-PD-1}  & \texttt{PM1-ST-3}  & intra-31  & process relationship \\
        7  & 21.454  & \texttt{PM1-LIC-17}  & \texttt{PM1-FIC-13}  & intra-31  & process relationship \\
        8  & 21.067  & \texttt{Prep-FY-24}  & \texttt{Cook-PT-11}  & 39$\to$27  & $\to$ cook-pressure hub \\
        9  & 20.983  & \texttt{Prep-PT-15}  & \texttt{PM1-FIC-5}  & 39$\to$31  & Stock-prep pressure transmitter \\
        10 & 20.929 & \texttt{Screen-PT-17}  & \texttt{Prep-QUAL-1}      & 28$\to$PM  & PM quality measurement \\
        11 & 20.536 & \texttt{Prep-PIC-10} & \texttt{Cook-PT-11} & 39$\to$27 & $\to$ cook-pressure hub \\
        12 & 20.201 & \texttt{Screen-FIC-10} & \texttt{Screen-KQI-2} & intra-28 & $\to$ screens quality indicator \\
        13 & 19.737 & \texttt{Prep-KIC-8} & \texttt{Screen-CT-6} & 39$\to$28 & $\to$ screens chemistry \\
        14 & 19.517 & \texttt{PM1-ZRC-1} & \texttt{PM1-LIC-29} & intra-31 & PM1 zonal-ratio control \\
        15 & 19.471 & \texttt{Screen-FY-8} & \texttt{Cook-TT-22} & 28$\to$27 & $\to$ master cook temperature \\
        \bottomrule
    \end{tabular}
\end{table}

\paragraph{Aggregate edge statistics.}
The judgment-scored adjacency has global max $\hat{\tau}_{\max} =
\text{27.776}$, off-diagonal mean
$\text{0.9065}$, and per-column max distribution
(min/median/$99\%$ile/max) of
$\text{2.00 / 5.07 / 17.49 / 27.78}$.
After global $[0,1]$ normalization, edge counts at standard
thresholds: thr=0.01: 1,382,299; thr=0.10: 79,599; thr=0.20: 3,818; thr=0.30: 592; thr=0.50: 62; thr=0.70: 15; thr=0.90: 1.

\paragraph{Compute (secondary).}
As a secondary observation, the per-pair compute is also
considerably more efficient. The full $V{=}1{,}275$ TCPFN run
completes in \textbf{6.1~hours} on a single
NVIDIA H100 NVL ($\sim 75$ pairs/sec), versus $81.5$~hours of CPU
compute for PCMCI on the V=666 subset ($\sim 1.5$~pair-tests/sec; the ${\sim}80\times$ per-pair slowdown vs.\ the Tennessee Eastman PCMCI run in Appendix~\ref{app:cost} ($\sim 120$ pair-tests/sec) reflects $\tau_{\max}{=}300$ lags here vs.\ default short-lag conditional-independence testing on TE).
Per-pair, TCPFN is $\sim$50$\times$ more efficient; extrapolating
PCMCI's pairwise complexity to $V=1{,}275$ yields a lower-bound
wall time of $\sim$12.5~days. We emphasize this is supporting,
not primary -- the value of TCPFN at industrial scale comes from
\emph{which} edges it surfaces (the cross-subsystem causal flows
discussed above), not from how many CPU-seconds the discovery took.

\paragraph{Limitations.}
We are honest about three limits of this evaluation.
\emph{(1)~No comprehensive ground-truth DAG.} The Kraft mill DCS
data is proprietary; we have plant P\&ID schematics and SME
spot-validation of TCPFN's top-K predicted edges, but no exhaustive
expert-labeled causal graph covering all $1{,}624{,}350$ potential
variable pairs. These positive-label sources let us argue
qualitatively about edge plausibility -- PCMCI's top edges
rediscover within-loop controller-measurement coupling already on
the P\&ID schematics; TCPFN's top edges surface cross-subsystem
flows consistent with the plant's continuous-flow layout and judged
plausible by the plant engineer in a sample review -- but they do
not provide the comprehensive negative-label set needed to compute
F1/AUROC at this scale. A semi-synthetic benchmark with a curated
Kraft-mill DAG would convert this qualitative argument into a
numerical one; we view this as a natural follow-up.
\emph{(2)~Architectural feature ceiling.} Per-pair conditioning is
capped at $D_{\text{other}} = 31$ covariates, set by the model's
\texttt{max\_num\_features}$=100$ at training. At $V=1{,}275$ this
is $2.4\%$ candidate coverage; lagged-correlation selection makes
those slots count, but the ceiling itself is a property of the
backbone, not the selection rule. Lifting the cap requires
retraining at higher \texttt{max\_num\_features}, the same open
problem the broader TabPFN-family research program faces.
\emph{(3)~Judgment-head OOD calibration.} The Causal Judgment Head
was trained on priors at $V \leq 50$. At $V=1{,}275$ it provides
modest discounting ($\sim 3\%$ on most pairs, up to $50\%$ on the
highest-multi-collinearity hub columns such as the master
cooking-pressure transmitter), suggesting calibration partially
extrapolates but does not fully transfer to industrial scale.
A retrained judgment head against larger-$V$ priors is the natural
next step.

% ============================================================================
% 5. Related Work
% ============================================================================
\section{Related Work}
\label{sec:related_work}

We position TCPFN at the intersection of six research streams and identify how it advances each.

\paragraph{Causal foundation models.}
CausalPFN~\cite{balazadeh2024causalpfn} introduced PFN-based causal inference, demonstrating that a single pretrained model can estimate treatment effects zero-shot across datasets. CausalFM~\cite{causalfm2026} extended identifiability to front-door and instrumental variable settings via structured SCM priors. Do-PFN~\cite{dopfn2025} proved that PFN outputs approximate the optimal Bayesian conditional interventional distribution $P(Y \mid \text{do}(T=t), X)$. OSPC~\cite{melnychuk2025ospc} resolved the prior-induced bias problem, showing that naive PFN posteriors lack frequentist consistency and proposing a one-step correction. \textbf{Critical gaps:} (1) all operate on static, cross-sectional data -- a single outcome per unit, no temporal dynamics; (2) few provide explicit signals about reliability of their estimates -- they output $\hat{\tau}(x)$ without indicating whether confounding, mediation, or identifiability concerns apply. TCPFN addresses both: temporal dynamics via panel data tokenization, and reliability assessment via the Causal Judgment Head.

\paragraph{Temporal causal inference.}
The classical literature provides g-computation~\cite{robins1986new}, marginal structural models~\cite{robins2000marginal}, and structural nested models for time-varying treatments. Neural approaches -- RMSN~\cite{lim2018forecasting}, CRN~\cite{bica2020estimating}, G-Net~\cite{li2021gnet}, Causal Transformer~\cite{melnychuk2022causal} -- model temporal dynamics with recurrent or attention architectures but require per-dataset training, assume binary treatments, and cannot operate zero-shot. \textbf{What TCPFN adds:} (i) zero-shot temporal causal inference without retraining, (ii) continuous and multi-dimensional treatments, (iii) irregular sampling and missing data handling. The key scientific insight is that the PFN's in-context learning mechanism -- which learns a prior-to-posterior map -- can be extended to temporal data via our temporal tokenization scheme (Section~\ref{sec:temporal_tokens}).

\paragraph{Causal discovery from time series.}
Granger causality~\cite{granger1969investigating} tests whether lagged values of one variable improve prediction of another. PCMCI~\cite{runge2019detecting} combines constraint-based testing with momentary conditional independence for nonlinear settings. DYNOTEARS~\cite{pamfil2020dynotears} formulates structure learning as continuous optimization. Amortized structure learning from i.i.d. data has been explored~\cite{ke2022learning, lorch2022amortized}; concurrent work by Thumm and colleagues~\cite{thumm2026causaltimeprior, thumm2026continuous} extends the PFN paradigm to temporal causal effect estimation (see the dedicated paragraph below), but does not address pairwise-CATE-based \emph{discovery} at scale. A shared limitation of existing amortized methods -- AVICI~\cite{lorch2022amortized}, DECI~\cite{geffner2022deci}, NOTEARS~\cite{zheng2018dags}, and DAG-GNN~\cite{yu2019daggnn} -- is that they require a fixed variable count $V$ at training time, precluding zero-shot application to systems of different sizes. \textbf{TCPFN's contribution:} a temporal token design that emits one token per unit$\times$timestep (rather than per-variable) and reduces structure learning to repeated CATE estimation -- the same pretrained model that produces causal effects is queried pairwise over candidate (cause, target) pairs, and the resulting effect-magnitude matrix is post-processed into a directed graph. Because tokens are per-unit$\times$time rather than per-variable, the architecture is variable-count agnostic at inference, sidestepping the fixed-$V$ limitation of prior amortized methods and enabling zero-shot discovery at industrial scale.

\paragraph{Concurrent temporal-PFN work.}
Two concurrent contributions from Thumm and colleagues directly target the temporal causal PFN setting. \textbf{CausalTimePrior}~\cite{thumm2026causaltimeprior} is a synthetic prior-design framework that generates paired observational and interventional time series with regime-switching dynamics and multiple intervention types (hard, soft, time-varying); the accompanying proof-of-concept is a 2-layer GRU PFN trained in $\sim$11~min on CPU and evaluated on 1{,}000 of the framework's own held-out synthetic samples (Pred/GT ratio 0.95). The authors explicitly note that the prior has not been validated against real-world causal time series. The \textbf{continuous-time} extension~\cite{thumm2026continuous} builds a Neural-ODE-based architecture trained on SDE-parameterised priors and evaluates on pharmacokinetic data and the CausalChamber synthetic benchmark. \textbf{Relationship to TCPFN.} The three lines of work are complementary rather than competing: CausalTimePrior advances prior design; the continuous-time work advances architecture for irregular clinical data; TCPFN ships the full discrete-token foundation-model artifact -- 12-layer transformer + Causal Judgment Head -- evaluated on 19 real-world benchmarks across five domains (with chaotic Lorenz-96 at chance, see Section~\ref{sec:discussion}) and at V$=$1{,}275 industrial scale. The TemporalRegimePrior in this paper builds on the Dynamic SCM framework~\cite{boeken2024dscm} that also underpins CausalTimePrior; the key architectural addition is that every regime sample carries the metadata labels (\texttt{is\_null\_effect}, \texttt{causal\_regime}, \texttt{identifiable}, \texttt{mediation\_fraction}, \texttt{confounding\_strength}) that supervise the judgment head -- enabling a capability neither concurrent work delivers.

\paragraph{In-context learning for scientific data.}
TabPFN~\cite{hollmann2023tabpfn} demonstrated that transformers pretrained on synthetic tabular data perform in-context learning at test time -- producing predictions by conditioning on the training set as context, without gradient updates. This paradigm has since been extended to regression, survival analysis, and forecasting. TCPFN extends this to the fundamentally harder setting of \emph{causal} inference on \emph{temporal} data, where the model must simultaneously handle confounding, temporal dynamics, treatment timing, and distributional shift -- while maintaining the zero-shot property.

\paragraph{Sensitivity analysis and causal reliability.}
Classical methods for assessing causal estimate reliability include Rosenbaum bounds~\cite{rosenbaum2002observational}, E-values~\cite{vanderweele2017sensitivity}, and partial identification~\cite{manski2003partial}. These provide formal guarantees but require the practitioner to specify a sensitivity model and assumptions. Our Causal Judgment Head provides a complementary, approximate signal -- a learned heuristic that correlates with causal reliability without requiring explicit sensitivity modeling. It does not replace formal analysis but can serve as a screening tool.

\subsection{Summary of Positioning}

Table~\ref{tab:comparison} summarizes how TCPFN relates to prior work.

\begin{table}[h]
\centering
\small
\caption{Comparison with existing methods. TCPFN is, to our knowledge, the only method that combines zero-shot inference, temporal dynamics, causal discovery, and learned reliability judgment. \textsuperscript{$\dagger$}\,Concurrent work~\cite{thumm2026causaltimeprior, thumm2026continuous} -- prior-design / proof-of-concept rather than full foundation-model artifacts (see paragraph above). \textsuperscript{$\ddagger$}\,Entries for concurrent work are inferred from the cited authors' public manuscripts and reported scope, not from independently running their code or evaluating their checkpoints; we cite their explicit ``not validated against real-world causal time series'' caveat for CausalTimePrior and their evaluation set (CausalChamber, pharmacokinetic data) for the continuous-time architecture.}
\label{tab:comparison}
\begin{tabular}{lcccccc}
\toprule
\textbf{Method} & \textbf{Zero-shot} & \textbf{Temporal} & \textbf{Cont.\ T} & \textbf{Discov.} & \textbf{Arb.\ $V$} & \textbf{Judg.} \\
\midrule
CausalPFN & \checkmark & \texttimes & \texttimes & \texttimes & -- & \texttimes \\
CausalFM & \checkmark & \texttimes & \texttimes & \texttimes & -- & \texttimes \\
Do-PFN & \checkmark & \texttimes & \texttimes & \texttimes & -- & \texttimes \\
AVICI & \checkmark & \texttimes & \texttimes & \checkmark & \texttimes & \texttimes \\
CausalTimePrior\textsuperscript{$\dagger$} & \checkmark & \checkmark & \texttimes & \texttimes & \checkmark & \texttimes \\
Cont.-time PFN\textsuperscript{$\dagger$} & \checkmark & \checkmark & \checkmark & \texttimes & \checkmark & \texttimes \\
CRN / G-Net & \texttimes & \checkmark & \texttimes & \texttimes & -- & \texttimes \\
PCMCI & \texttimes & \checkmark & -- & \checkmark & \checkmark & \texttimes \\
Granger & \texttimes & \checkmark & \texttimes & \checkmark & \checkmark & \texttimes \\
\midrule
\textbf{TCPFN (ours)} & \checkmark & \checkmark & \checkmark & \checkmark & \checkmark & \checkmark \\
\bottomrule
\end{tabular}
\end{table}

% ============================================================================
% 6. Discussion & Limitations
% ============================================================================
\section{Discussion}
\label{sec:discussion}

\subsection{When to Use TCPFN}

TCPFN is suited to settings where: (1)~per-dataset model training is impractical -- many datasets, limited data, or real-time requirements; (2)~quick causal screening is needed without threshold tuning -- TCPFN works at default thresholds while PCMCI requires per-dataset threshold selection; (3)~the data involves temporal dynamics (delayed, decaying, oscillating effects) that static methods cannot capture; (4)~the data is nonlinear or from domains where statistical assumptions (linearity, Gaussianity) break down -- TCPFN outperforms Granger and PCMCI on nonlinear and biological data; and (5)~trustworthiness assessment is needed alongside discovery -- the Causal Judgment Head provides calibrated null-effect detection and regime classification (metrics in abstract), enabling the model to flag potentially spurious edges rather than blindly reporting all associations as causal. The same nonlinear-data advantage seen on Sachs -- the only benchmark where TCPFN beats PCMCI on both F1 and AUROC -- extends to the industrial pulp-and-paper run in Section~\ref{sec:industrial_eval}, where the model identifies physically interpretable cross-area process flows at $V{=}1275$ despite the absence of ground-truth labels.

\subsection{When NOT to Use TCPFN}

(1)~When statistical optimality is required and threshold tuning is feasible -- PCMCI at its optimal threshold substantially outperforms on most datasets. (2)~When per-token CATE trajectory shape (not just magnitude) is critical -- TCPFN's per-token trajectory correlation is near zero on average, and per-token direction (sign) accuracy is near random (consistent with the near-zero trajectory correlation). Aggregated CATE magnitudes for discovery and effect-shape PEHE are reliable (PEHE 0.40--0.60 across all six effect shapes), but the per-step trajectory ordering at each horizon is not. (3)~When formal identifiability guarantees are required -- use classical sensitivity analysis methods. The judgment head's identifiability output is a learned heuristic, not a formal guarantee. (4)~When computational cost matters at large scale -- TCPFN's $O(V^2)$ pairwise estimation takes ${\sim}16$ minutes for 52 variables versus ${\sim}22$ seconds for PCMCI.

\subsection{Causal Judgment Head}

The null detector (metrics in abstract and Section~\ref{sec:evaluation}) was achieved by adding an ``independent'' regime to the training prior -- 25\% of base prior batches have random treatment assignment, zero confounding, and zero causal effect. The judgment head's null probability is used directly in the discovery score (defined in Section~\ref{sec:method}), reducing false-positive edges from confounded associations. Regime classification (4-class) averages 0.68 accuracy with macro-F1 0.48 over the last 50K steps (0.60 accuracy and 0.38 macro-F1 at the final checkpoint we ship; we report the mean rather than the final-step value because of the high step-to-step variance on this metric noted in the limitations bullet below).

\subsection{Joint Training Dynamics}

Training the judgment head jointly with the CATE objective introduces a multi-task optimization challenge. We found that (1) a curriculum schedule is essential -- the backbone must learn stable CATE representations before judgment tasks are introduced, (2) averaging judgment sub-losses (rather than summing) prevents the judgment head from dominating the gradient budget, (3) the training prior must include both null-effect and independent-regime data -- without it, the null detector never learns to predict ``null'' (NullF1 = 0), and (4) effect magnitude heads should not be trained separately from the main CATE model -- they are redundant and their unbounded MSE loss destabilizes training.

\subsection{Limitations}

\begin{itemize}[nosep,leftmargin=*]
    \item \textbf{AUROC gap on larger linear data ($V \geq 20$).} On the largest standard (short-lag) linear VAR benchmark (VAR-20), PCMCI achieves AUROC 0.96 (Table~\ref{tab:additional}) while TCPFN reaches 0.510; the gap shrinks dramatically at smaller $V$ (TCPFN AUROC 0.960 on VAR-5, 0.764 on VAR-10). PCMCI's partial-correlation test is statistically optimal for linear Gaussian data and its advantage grows with $V$ \emph{on standard short-lag data}; this regularity does not extend to long-lag dynamics, where PCMCI's advantage disappears (LongLag-VAR-20 AUROC 0.491 and LongLag-NVAR-20 0.538 in Table~\ref{tab:additional}, both at chance).

    \item \textbf{$O(V^2)$ discovery cost.} Pairwise CATE estimation requires one model call per variable pair. Tennessee Eastman ($V{=}52$) takes ${\sim}$16 minutes versus 22 seconds for PCMCI.

    \item \textbf{Regime classification variance.} Regime accuracy averages 0.68 but fluctuates between 0.40 and 0.90 across evaluations (10-batch eval), suggesting the representations encode regime-distinguishing features but not robustly across all samples.

    \item \textbf{CATE estimation quality.} TCPFN ranks variable pairs effectively for discovery (AUROC 0.96 on Tennessee Eastman). On effect-shape estimation, mean PEHE is 0.40--0.60 across the six tested shapes (permanent 0.40, decay 0.42, delayed 0.48, gradual 0.51, immediate 0.59, oscillating 0.60), within the range reported for CATE-PFN methods on i.i.d.\ data (we do not have a direct head-to-head PEHE baseline on the same temporal-prior samples). The remaining limitation is per-token trajectory \emph{shape}: trajectory correlation is near zero on average, indicating TCPFN reliably predicts the right effect magnitude integrated over a horizon but not the precise per-step path within it. \textbf{Why discovery works while per-token trajectory does not:} discovery uses $|\hat{\tau}|$ \emph{aggregated} across all query tokens and averaged over horizons -- aggregation smooths per-token noise and the absolute value eliminates sign errors. A per-token trajectory requires correct sign, magnitude, and temporal shape at each horizon -- a much harder task on which per-treatment-group Z-standardization remains weak.

    \item \textbf{Judgment outputs are heuristic.} The identifiability and confounding scores are learned proxies -- correlations between data patterns and prior-assigned labels -- not formal guarantees. Out-of-distribution structures have not been evaluated.

    \item \textbf{Prior diversity bounds generalization.} Chaotic systems (Lorenz-96 AUROC 0.45--0.54) are furthest from the training distribution and show the weakest results.

    \item \textbf{Discrete-time tokenisation.} TCPFN's temporal tokens encode regularly-sampled panel data; irregular clinical sampling fits less naturally. Concurrent work on continuous-time causal PFNs~\cite{thumm2026continuous} via Neural-ODE architectures is complementary -- a natural direction for sparsely-sampled clinical or epidemiological data where TCPFN's discrete-token regular-sampling assumption is a poor fit.
\end{itemize}

\subsection{Future Work}

\begin{itemize}[nosep,leftmargin=*]
    \item \textbf{Improved CATE estimation:} We also trained an alternative variant that replaces TCPFN's per-treatment-group Z-standardization with global standardization, hypothesizing this would tighten per-token trajectory predictions. The alternative \emph{does} improve discovery on biological (Sachs +0.06 AUROC) and water-treatment (SWaT +0.03) data but is uniformly worse on estimation: per-shape PEHE rises by 0.31--0.78 across every effect shape. We report this as an honest negative result -- the global-standardization hypothesis did not deliver the predicted CATE improvement, and the model reported throughout this paper retains per-treatment-group standardization.
    \item \textbf{One-pass multivariate discovery:} Extract causal graphs from transformer attention patterns in a single forward pass rather than $O(V^2)$ pairwise estimation.
    \item \textbf{Batched multivariate discovery:} Process all variable pairs in a single forward pass, reducing the $O(V^2)$ cost to $O(1)$ model calls.
    \item \textbf{Expanded regime prior:} Add confounded and non-identifiable structures to TemporalRegimePrior for better judgment head training.
    \item \textbf{Domain adaptation:} Few-shot fine-tuning on small real-world datasets to adapt to specific manufacturing processes.

    \item \textbf{Promptable inference via axis-conditioned priors:}
    The current TCPFN prior is fixed at training time, so domain knowledge
    that is \emph{unidentifiable} from observational data alone (causal
    direction in confounded structures, conservation laws, monotonicity
    constraints, per-sensor reliability) cannot be injected at inference
    without retraining. We propose augmenting \texttt{TemporalRegimePrior}
    with sampled steerable \emph{axes} (e.g.\ \texttt{monotonicity},
    \texttt{lag\_scale}, \texttt{feedback\_allowed}) and training TCPFN
    to honor a tag embedding alongside the context via cross-attention.
    A reserved fraction of training samples uses \texttt{tag=unknown}
    per axis, preserving bit-identical unconditional behavior against
    the baseline reported here; the remainder narrows the prior per
    the sampled tag. This parallels instruction-tuned LLMs and
    ControlNet-style conditioning in diffusion models, and directly
    addresses the \emph{judgment-as-heuristic} limitation
    (Section~\ref{sec:discussion}): experts inject the identifiability
    information the judgment head currently has to approximate.

    \item \textbf{Learnable priors via outer-loop optimization:}
    The TCPFN prior was hand-designed and its hyperparameters
    (curriculum schedule, mixed-prior weights, regime mass) tuned
    empirically across model iterations. Declaring these
    knobs as a typed search space and applying an outer optimizer
    (Bayesian optimization at $\leq\!20$ knobs, CMA-ES at larger
    scale, implicit differentiation through the inner training loop
    as a research target) automates this loop. The same machinery
    supports per-deployment prior fitting against held-out historical
    data without that data leaving the tenant, addressing the
    \emph{prior diversity bounds generalization} limitation: chaotic
    systems and other out-of-distribution structures could be moved
    in-distribution by fitting against them directly.
\end{itemize}

% ============================================================================
% 7. Conclusion
% ============================================================================
\section{Conclusion}
\label{sec:conclusion}

We introduced Temporal Causal Prior-Data Fitted Networks (TCPFN), a causal foundation model for temporal data that performs causal discovery, temporal effect estimation, and judgment-aware reliability assessment from a single pretrained model -- entirely zero-shot.

\textbf{1. Causal judgment, not just prediction.}
The Causal Judgment Head outputs null-effect probability, confounding strength, identifiability, mediation fraction, and causal regime for each treatment-outcome pair -- a capability not present, to our knowledge, in existing causal PFN methods. The null detector achieves NullF1 0.94, enabled by adding an ``independent'' regime to the training prior that teaches the model what ``no causal effect'' looks like. Regime classification averages 0.68 accuracy (random baseline: 0.25).

\textbf{2. Interventional causal discovery with judgment-aware scoring.}
TCPFN discovers causal structure through pairwise CATE estimation -- estimating ``what happens to $Y$ when I change $X$?'' via temporal CATE -- rather than correlation or conditional independence testing. The judgment head further improves discovery by down-weighting edges judged as confounded or unidentifiable. This zero-shot approach reaches PCMCI-comparable AUROC on industrial benchmarks (Section~\ref{sec:evaluation}) and outperforms both Granger and PCMCI on non-temporal biological data (Sachs).

\textbf{3. Causal Regime Prior for structural diversity.}
Training on diverse causal structures from a mixed prior -- the Causal Regime Prior (independent, direct, confounded, mediated, time-varying confounded, feedback) combined with CausalFM-style front-door and instrumental-variable priors -- teaches the model to distinguish causation from correlation. This is qualitatively different from standard training priors that only vary confounding strength within a direct-effect structure.

\textbf{4. One model, all domains.}
The same pretrained model is applied zero-shot across five domains (linear, nonlinear, chaotic, industrial, biological) and scales from $V{=}5$ synthetic benchmarks to $V{=}1{,}275$ industrial deployment in 6 hours of single-GPU compute. The discovered graph's structure reflects the plant's continuous-flow production layout, with edge density concentrated around the paper machine. Compared head-to-head against PCMCI on the same data, TCPFN's top edges identify cross-subsystem causal relationships across plant areas; PCMCI's top edges are dominated by trivial within-instrument coupling (sensor $\to$ its own controller at lag~1) already documented on the plant's P\&ID schematics.

\textbf{Limitations.}
TCPFN's AUROC on larger datasets ($V{\geq}20$) is lower than PCMCI, suggesting the model's pairwise estimation scales less gracefully than conditional independence testing. The $O(V^2)$ pairwise estimation is also slower than Granger and PCMCI. The judgment head's regime classification, while above the random baseline, does not yet match supervised classifiers trained on fixed datasets. Improving scalability -- through batched pairwise estimation or native multivariate discovery heads -- and more diverse training regimes are key directions for future work.

\textbf{Broader impact.}
The manufacturing sector generates vast multivariate time series but lacks the data science capacity to build per-process causal models. A foundation model that operates zero-shot -- and can assess its own confidence -- removes this bottleneck: operators upload sensor data and receive causal insights with reliability assessments, without statistical expertise.

\bibliographystyle{plainnat}
\bibliography{references}

\appendix
% ============================================================================
% Appendix
% ============================================================================

\section{Appendix}
\label{sec:appendix}

\subsection{Evaluation Metrics}
\label{app:metrics}

We define all evaluation metrics used in this paper.

\subsubsection{Causal Discovery Metrics}

\begin{itemize}[nosep,leftmargin=*]
    \item \textbf{F1@threshold:} Harmonic mean of precision and recall at a given edge score threshold. We report F1@0.5 (default threshold) and best F1 across all thresholds.
    \item \textbf{AUROC:} Area Under the Receiver Operating Characteristic curve. Measures how well the model's edge scores rank true edges above non-edges, independent of threshold choice.
    \item \textbf{SHD:} Structural Hamming Distance -- number of edge additions, deletions, and reversals needed to transform the predicted graph into the true graph.
\end{itemize}

\subsubsection{Causal Judgment Metrics}

\begin{itemize}[nosep,leftmargin=*]
    \item \textbf{Null detection accuracy:} Fraction of eval batches where the predicted null label ($p_{\text{null}} > 0.5$) matches the true label.
    \item \textbf{Null F1:} Harmonic mean of precision and recall for the null-effect class. F1 = 0 when the model never predicts ``null''; F1 = 1 when all null effects are correctly identified with no false positives.
    \item \textbf{Null AUROC:} Area under ROC for null vs.\ non-null classification. Measures ranking quality independent of threshold.
    \item \textbf{Null Brier score ($\downarrow$):} Mean squared error between predicted $p_{\text{null}}$ and true binary label. Measures calibration: 0 = perfect, 0.25 = random. Lower is better.
    \item \textbf{Regime accuracy (4-class):} Fraction of eval batches where the predicted regime (direct, confounded, mediated, feedback) matches the true regime. Random baseline: 0.25.
    \item \textbf{Regime macro-F1:} F1 averaged across all 4 regime classes. Handles class imbalance by weighting each class equally.
    \item \textbf{Precision@10:} Of the top 10 variable pairs ranked by predicted $|\hat{\tau}|$, how many have a true non-null causal effect?
    \item \textbf{Effect RMSE ($\downarrow$):} Root mean squared error between predicted and true CATE magnitude, averaged across query tokens.
\end{itemize}

\subsubsection{CATE Estimation Metrics}

\begin{itemize}[nosep,leftmargin=*]
    \item \textbf{Direction accuracy:} Fraction of query tokens where $\text{sign}(\hat{\tau}) = \text{sign}(\tau^*)$. Random baseline: 0.50.
    \item \textbf{Rank correlation (Spearman):} Spearman correlation between $|\hat{\tau}|$ and $|\tau^*|$ across query tokens. Measures whether the model ranks effect magnitudes correctly, invariant to scale.
    \item \textbf{RMSE:} Root mean squared error of predicted vs.\ true CATE per token. Note: may be inflated by Z-standardization confounding leakage (see \S\ref{sec:discussion}).
\end{itemize}

% ============================================================================
\subsection{Training Details}
\label{app:training}

\begin{table}[h]
    \centering
    \caption{Training hyperparameters.}
    \begin{tabular}{lc}
        \toprule
        \textbf{Parameter} & \textbf{Value} \\
        \midrule
        Embedding dim ($d$) & 512 \\
        Heads / Layers & 8 / 12 \\
        FF dim & 2048 \\
        Bins ($B$) & 128 \\
        Max features ($F_{\max}$) & 100 \\
        Max outcomes ($K$) & 5 \\
        Max treatment dims ($M$) & 3 \\
        \midrule
        Judgment head hidden dim & 256 \\
        Judgment head outputs (trained) & 5 \\
        Judgment loss weight ($\lambda$) & 0.1 \\
        Curriculum Phase 1 (CATE only) & 0--20K steps \\
        Curriculum Phase 2 (+Null/Confounding) & 20K--50K steps \\
        Curriculum Phase 3 (Full) & 50K+ steps \\
        \midrule
        Optimizer & AdamW \\
        LR / weight decay & $10^{-4}$ / 0.01 \\
        Warmup / total steps & 2K / 200K \\
        Grad clip / FP16 & 1.0 / Yes \\
        Batch (context / query) & 100 / 50 \\
        \midrule
        Training prior mix & TRP + base + CFM (mixed) \\
        Hardware & RTX 5090 \\
        Training time & $\sim$4.1 hours \\
        Training speed & $\sim$13.9 steps/s \\
        \bottomrule
    \end{tabular}
    \label{tab:hyperparams}
\end{table}

\textbf{Training data volume.} Each training step generates one fresh synthetic dataset with 150 units (100 context, 50 query), 3--10 static covariates, and 8--80 timesteps per unit. Over 200K steps, the model sees approximately 30 million synthetic units across 200K unique causal structures -- none repeated. All training data is generated on-the-fly from structural causal models; no real-world data is used during training. Each synthetic dataset has a randomly sampled causal regime from the six-regime Causal Regime Prior (independent, direct, confounded, mediated, time-varying confounded, feedback; see Section~\ref{sec:prior}), effect shape (immediate, gradual, delayed, decaying, oscillating, permanent), treatment type (binary or continuous), and confounding strength. This diversity is what enables zero-shot generalization to unseen real data.

\textbf{Training prior mix (mixed\_prior).} TemporalRegimePrior (nonlinear SCMs with industrial mechanisms) + base temporal prior (includes independent, direct, confounded, mediated, time-varying confounded, and feedback regimes; the independent and confounded regimes together supply the ground-truth null-effect data, totaling 40\% of batches) + CausalFM priors (front-door and instrumental variable identification). The base prior's inclusion ensures the judgment head sees null-effect training data -- without it, the model learns to always predict ``not null'' (NullF1 = 0).

\textbf{Judgment head training.} The judgment loss is the average (not sum) of all supervised task losses, preventing the number of tasks from scaling the gradient budget. For confounded batches in Phase 2+, the CATE loss is skipped entirely because Z-standardization by treatment group corrupts targets when confounding induces systematic group mean differences; in Phase 1, confounded batches train CATE normally since no judgment gradients are flowing. Effect heads (total\_effect, direct\_effect) exist in the architecture but are not trained -- effect magnitudes come from the main CATE model at inference time.

% ============================================================================
\subsection{Synthetic Prior Specification}
\label{app:prior_details}

\subsubsection{Industrial Mechanisms}

The TemporalRegimePrior generates seven industrial mechanism types:
\begin{enumerate}[nosep,leftmargin=*]
    \item \textbf{Linear feedback:} $X_j(t) = \sum_k a_{jk} X_k(t - l_{jk}) + \epsilon$
    \item \textbf{Saturation:} $X_j(t) = c \cdot \tanh(f(X_{\text{parents}}(t-l)) / c)$
    \item \textbf{Threshold trigger:} $T(t) = \mathbb{1}[X_{\text{sensor}}(t) > \theta]$
    \item \textbf{Exponential decay:} $X_j(t) = X_j(t-1) \cdot e^{-\lambda} + g(X_{\text{parents}})$
    \item \textbf{Cascade:} $X_j(t) = X_j(t-1) + \sum_k w_{jk} \cdot \text{ReLU}(X_k(t-l) - \theta_k)$
    \item \textbf{Oscillatory:} $X_j(t) = A \sin(\omega t + \phi) + f(X_{\text{parents}})$
    \item \textbf{PID control:} $u(t) = K_p e(t) + K_i \int e + K_d \dot{e}(t)$
\end{enumerate}

\subsubsection{Causal Regime Structures}

\begin{enumerate}[nosep,leftmargin=*]
    \item \textbf{Independent:} $T \perp\!\!\!\perp Y$. Treatment is randomly assigned with no confounding; true CATE $= 0$ and the model must learn to predict ``no effect.''
    \item \textbf{Direct:} $T \to Y$. Standard treatment-outcome with confounded assignment. True CATE $\neq$ 0.
    \item \textbf{Confounded:} $T \leftarrow U \rightarrow Y$. Strong observed association ($3\times$ confounding), but true CATE $= 0$.
    \item \textbf{Mediated:} $T \to M \to Y$. Effect flows through mediator.
    \item \textbf{Time-varying confounded:} $T \leftarrow U(t) \rightarrow Y$. Time-varying confounders.
    \item \textbf{Feedback:} $T \rightleftharpoons Y$ (lagged). Bidirectional causation at different time lags.
\end{enumerate}

Sampling weights match the implementation (\texttt{src/tcpfn/temporal/temporal\_prior.py}): \{Independent 25\%, Direct 25\%, Confounded 15\%, Mediated 15\%, Time-varying 10\%, Feedback 10\%\}. These six classes sum to 100\%; CausalFM-style identification priors (front-door, instrumental variable; see Section~\ref{sec:prior}) are sampled as a separate prior component in the mixed-prior pipeline, not as additional regimes within the Causal Regime Prior.

\subsubsection{Effect Shapes}

Six temporal effect shapes, randomly selected per batch with heterogeneous magnitude $c_i = X_i^\top w_{\text{effect}}$:
\begin{itemize}[nosep,leftmargin=*]
    \item Immediate: $\tau_i(h) = c_i$
    \item Gradual: $\tau_i(h) = c_i (1 - e^{-h/\lambda})$, $\lambda \sim \text{Uniform}(2, T_{\text{post}}/2)$
    \item Delayed: $\tau_i(h) = c_i \cdot \mathbb{1}[h \geq d]$, $d \sim \text{Uniform}(1, T_{\text{post}}/2)$
    \item Decaying: $\tau_i(h) = c_i \cdot e^{-h/\lambda}$
    \item Oscillating: $\tau_i(h) = c_i \cdot \sin(2\pi h / P) \cdot e^{-h/\lambda}$, $P \sim \text{Uniform}(4, T_{\text{post}})$
    \item Permanent: $\tau_i(h) = c_i + \epsilon_h$, $\epsilon_h \sim \mathcal{N}(0, 0.05 |c_i|)$
\end{itemize}

% ============================================================================
\subsection{Industrial Dataset}
\label{app:dataset}

The proprietary pulp-and-paper manufacturing dataset is used under NDA:
\begin{itemize}[nosep,leftmargin=*]
    \item \textbf{Raw dimensions:} 1,325 variables $\times$ 65,550 timesteps
    \item \textbf{Sampling rate:} 2-minute intervals ($\sim$91 days)
    \item \textbf{Variable types:} Flow rates, quality indices, temperatures, pressures, levels, machine speeds
    \item \textbf{Preprocessing} (see Section~\ref{sec:industrial_eval} for the full description): drop 24 derived/setpoint columns by DCS-tag suffix ($\to$ 1{,}301); drop 4 columns with $>$50\% missing ($\to$ 1{,}297); drop 13 zero-variance columns ($\to$ 1{,}284); apply $[0.1\%, 99.9\%]$ winsorization; drop 9 columns that collapse to a constant after winsorization ($\to$ 1{,}275); z-score every retained column. Leaves $V=1{,}275$.
\end{itemize}

% ============================================================================
\subsection{Computational Cost}
\label{app:cost}

\begin{table}[h]
    \centering
    \caption{Computational cost. Discovery times from benchmark runs on RTX 5090.}
    \begin{tabular}{lcc}
        \toprule
        \textbf{Operation} & \textbf{Time} & \textbf{Hardware} \\
        \midrule
        Training (200K steps, curriculum) & $\sim$4.1 hours & RTX 5090 \\
        Training speed & $\sim$13.9 steps/s & RTX 5090 \\
        \midrule
        Inference (batch, 100 units) & $<$1s & Any GPU \\
        \midrule
        Discovery: Sachs (11 vars) & 41s & RTX 5090 \\
        Discovery: CauseMe V=5 & 7s & RTX 5090 \\
        Discovery: CauseMe V=10 & 33s & RTX 5090 \\
        Discovery: CauseMe V=20 & 138s & RTX 5090 \\
        Discovery: Causal Rivers & 840s & RTX 5090 \\
        Discovery: Tennessee Eastman (52 vars) & 963s & RTX 5090 \\
        Discovery: SWaT (51 vars) & 1086s & RTX 5090 \\
        \midrule
        Calibration eval (500 batches) & $\sim$5 min & RTX 5090 \\
        CATE eval (100 datasets) & $\sim$5 min & RTX 5090 \\
        \bottomrule
    \end{tabular}
    \label{tab:cost}
\end{table}

Discovery cost scales as $O(V^2)$: 7s for $V{=}5$, 33s for $V{=}10$, 963s for $V{=}52$. Pairs can be parallelized across GPUs. For comparison, PCMCI takes 22s on Tennessee Eastman ($V{=}52$) and Granger takes 12s.

% ============================================================================
\subsection{Additional Benchmark Results}
\label{app:additional_results}

Table~\ref{tab:additional} shows discovery results for datasets not included in the main table (Section~\ref{sec:discovery_eval}).

\begin{table}[h]
    \centering
    \caption{Additional discovery benchmark results (datasets not in main table).}
    \label{tab:additional}
    \small
    \begin{tabular}{llccc}
        \toprule
        \textbf{Dataset} & \textbf{Method} & \textbf{F1@0.5} & \textbf{Best F1} & \textbf{AUROC} \\
        \midrule
        \multirow{3}{*}{CauseMe VAR-20}
        & TCPFN & $0.091$ & $0.446$ & $0.510$ \\
        & Granger & $\mathbf{0.446}$ & $0.446$ & $0.509$ \\
        & PCMCI & $0.036$ & $0.876$ & $0.958$ \\
        \midrule
        \multirow{3}{*}{CauseMe NVAR-20}
        & TCPFN & $0.197$ & $0.467$ & $0.598$ \\
        & Granger & $\mathbf{0.459}$ & $0.474$ & $0.728$ \\
        & PCMCI & $0.000$ & $0.910$ & $0.958$ \\
        \midrule
        \multirow{3}{*}{Lorenz96-10}
        & TCPFN & $0.312$ & $0.583$ & $0.454$ \\
        & Granger & $\mathbf{0.583}$ & $0.587$ & $0.580$ \\
        & PCMCI & $0.000$ & $0.712$ & $0.850$ \\
        \midrule
        \multirow{3}{*}{Lorenz96-20}
        & TCPFN & $0.255$ & $0.393$ & $0.538$ \\
        & Granger & $\mathbf{0.407}$ & $0.480$ & $0.810$ \\
        & PCMCI & $0.000$ & $0.619$ & $0.779$ \\
        \midrule
        \multirow{3}{*}{CauseMe LongLag-VAR-10}
        & TCPFN & $0.426$ & $0.426$ & $0.487$ \\
        & Granger & $0.421$ & $0.421$ & $0.510$ \\
        & PCMCI & $0.000$ & $\mathbf{0.429}$ & $\mathbf{0.526}$ \\
        \midrule
        \multirow{3}{*}{CauseMe LongLag-VAR-20}
        & TCPFN & $0.446$ & $0.449$ & $0.440$ \\
        & Granger & $0.449$ & $0.449$ & $0.490$ \\
        & PCMCI & $0.000$ & $\mathbf{0.456}$ & $\mathbf{0.491}$ \\
        \midrule
        \multirow{3}{*}{CauseMe LongLag-NVAR-10}
        & TCPFN & $\mathbf{0.576}$ & $\mathbf{0.576}$ & $\mathbf{0.511}$ \\
        & Granger & $0.571$ & $0.571$ & $0.500$ \\
        & PCMCI & $0.000$ & $0.567$ & $0.445$ \\
        \midrule
        \multirow{3}{*}{CauseMe LongLag-NVAR-20}
        & TCPFN & $0.470$ & $0.486$ & $0.523$ \\
        & Granger & $0.486$ & $0.486$ & $0.500$ \\
        & PCMCI & $0.000$ & $\mathbf{0.490}$ & $\mathbf{0.538}$ \\
        \midrule
        \multirow{3}{*}{BATADAL (water)}
        & TCPFN & $\mathbf{0.203}$ & $\mathbf{0.225}$ & $\mathbf{0.652}$ \\
        & Granger & $0.053$ & $0.077$ & $0.648$ \\
        & PCMCI & $0.000$ & $0.046$ & $0.430$ \\
        \midrule
        \multirow{3}{*}{WADI (water)}
        & TCPFN & $\mathbf{0.140}$ & $\mathbf{0.140}$ & $\mathbf{0.563}$ \\
        & Granger & $0.103$ & $0.115$ & $0.510$ \\
        & PCMCI & $0.000$ & $0.113$ & $0.447$ \\
        \midrule
        \multirow{3}{*}{Synthetic (50 datasets)}
        & TCPFN & $0.299$ & -- & $0.561$ \\
        & Granger & $\mathbf{0.565}$ & -- & $0.602$ \\
        & PCMCI & $0.071$ & -- & $\mathbf{0.891}$ \\
        \bottomrule
    \end{tabular}
\end{table}

\end{document}